%% file: acl2023.tex
\pdfoutput=1

\documentclass[11pt]{article}

\usepackage{ACL2023}

\usepackage{times}
\usepackage{latexsym}

\usepackage[T1]{fontenc}

\usepackage[utf8]{inputenc}

\usepackage{microtype}

\usepackage{inconsolata}

\input{00-header}
\title{\shortname: Natural Language Visual Reasoning with Reinforcement Learning}

\author{Anne Wu, Kianté Brantley, Noriyuki Kojima, and Yoav Artzi\\
Department of Computer Science and Cornell Tech, Cornell University\\
\texttt{\{aw588, kdb82, nk654\}@cornell.edu, \{yoav\}@cs.cornell.edu}}

\begin{document}

\maketitle

\begin{abstract}

\input{10-abstract.tex}

\end{abstract}

\input{20-intro.tex}

\input{25-related.tex}

\input{35-background.tex}

\input{30-task.tex}

\input{40-dataset.tex}

\input{50-exp.tex}

\input{80-conclusion.tex}

\input{85-limitations.tex}

\input{87-ethics.tex}

Commented for anonymous submission
\input{90-acks.tex}

\bibliography{custom}
\bibliographystyle{acl_natbib}

\clearpage

\appendix

\input{100-app_task.tex}

\input{110-app_dataset.tex}

\input{120-app_experimental_design.tex}

\input{130-exp.tex}

\input{132-other_models}

\input{135-app_third_party.tex}

\input{tables/71-ling-analysis_ppo_full.tex}
\input{tables/71-ling-analysis_ppo_sf_full.tex}

\end{document}

%% file: 00-header.tex
\usepackage{times}
\usepackage{latexsym}
\usepackage{graphicx}
\usepackage{float}
\usepackage{amsmath}
\usepackage[T1]{fontenc}
\usepackage{multirow}
\usepackage[utf8]{inputenc}
\usepackage{xspace}

\usepackage{booktabs}

\usepackage{hyperref}
\usepackage{url}

\usepackage{braket}
\usepackage{tikz}
\usepackage{pgfplots}
\usepackage{amssymb}
\usepackage{xcolor}
\usepackage{svg}
\usepackage{tikzscale}
\usepackage{graphicx}
\usepackage{subcaption}
\usepackage{longtable}
\usepackage{xcolor}
\usepackage{wrapfig}
\usepackage{nameref}
\usepackage{lscape, rotating}
\usepackage{enumitem}
\usepackage{listings}
\usepackage{afterpage}

\usepackage{ulem}

\definecolor{codegreen}{rgb}{0,0.6,0}
\definecolor{codegray}{rgb}{0.5,0.5,0.5}
\definecolor{codepurple}{rgb}{0.58,0,0.82}
\definecolor{backcolour}{rgb}{0.95,0.95,0.92}

\lstdefinestyle{mystyle}{
    commentstyle=\color{codegreen},
    keywordstyle=\color{magenta},
    numberstyle=\tiny\color{codegray},
    stringstyle=\color{codepurple},
    basicstyle=\ttfamily\footnotesize,
    breakatwhitespace=false,         
    breaklines=true,                 
    captionpos=b,                    
    keepspaces=true,                 
    numbersep=1pt,                  
    showspaces=false,                
    showstringspaces=false,
    showtabs=false,                  
    tabsize=2
}

\usepackage{arydshln} %
\usetikzlibrary{pgfplots.groupplots}
\usetikzlibrary{positioning}
\usetikzlibrary{fit}

\pgfplotsset{compat=1.10}
\usepgfplotslibrary{fillbetween}

\newcommand{\eat}[1]{}

\newcommand{\nlstring}[1]{\textit{#1}}
\newcommand{\pmstring}[1]{\scriptsize$\pm${#1}}

\newcommand{\aautoref}[1]{\hyperref[#1]{Appendix~\ref*{#1}}}

\newcommand{\envname}[1]{\texttt{\MakeUppercase{#1}}\xspace}
\newcommand{\envtower}{\envname{tower}}
\newcommand{\envscatter}{\envname{scatter}}

\newcommand{\mappingfunc}{\mathcal{M}}

\newcommand{\context}{c}
\newcommand{\contextset}{\mathcal{C}}

\newcommand{\action}{a}
\newcommand{\actionset}{\mathcal{A}}

\newcommand{\act}[1]{\mathtt{#1}}
\newcommand{\stopaction}{\act{STOP}}

\newcommand{\state}{s\xspace}

\newcommand{\stateset}{\mathcal{S}}
\newcommand{\initialstateset}{\beta^c}

\newcommand{\transitionfunc}{T}
\newcommand{\hor}{H\xspace}

\newcommand{\taskname}[1]{\texttt{\MakeUppercase{#1}}\xspace}
\newcommand{\taskscratch}{\taskname{scratch}}
\newcommand{\taskflipit}{\taskname{flipit}}

\newcommand{\rewardfunc}{R^c}

\newcommand{\token}{x}
\newcommand{\tokenseq}{\bar{\token}}

\newcommand{\logicalformfunc}{\mathcal{E}^{\tokenseq}}

\newcommand{\targetbool}{b}
\newcommand{\true}{\texttt{True}\xspace}
\newcommand{\false}{\texttt{False}\xspace}

\newcommand{\policy}{\pi}
\newcommand{\params}{\theta}

\newcommand{\real}{\mathbb{R}}

\newcommand{\shortname}{\textit{lil}Gym\xspace}

\newcommand{\nlvr}{NLVR\xspace}
\newcommand{\benchmarkurl}{\url{https://lil.nlp.cornell.edu/lilgym/}\xspace}

\newcommand{\cnnbase}{C3+BERT\xspace}
\newcommand{\vilt}{ViLT\xspace}
\newcommand{\flava}{FLAVA\xspace}
\newcommand{\clip}{CLIP\xspace}
\newcommand{\cnnbaseten}{C10+BERT\xspace}
\newcommand{\ppo}{PPO\xspace}
\newcommand{\ppowsf}{PPO+SF\xspace}

\newcommand{\tost}{\envname{Tower-Scratch\xspace}}
\newcommand{\toft}{\envname{Tower-FlipIt\xspace}}
\newcommand{\scst}{\envname{Scatter-Scratch\xspace}}
\newcommand{\scft}{\envname{Scatter-FlipIt\xspace}}

\definecolor{colortrue}{HTML}{d3273e}
\definecolor{colorfalse}{HTML}{d6d2c4}

%% file: 10-abstract.tex
We present \shortname, a new benchmark for language-conditioned reinforcement learning in visual environments. \shortname is based on 2{,}661 highly-compositional human-written natural language statements grounded in an interactive visual environment. We introduce a new approach for exact reward computation in every possible world state by annotating all statements with executable Python programs. 
Each statement is paired with multiple start states and reward functions to form thousands of distinct Markov Decision Processes of varying difficulty. 
We experiment with \shortname with different models and learning regimes. Our results and analysis show that while existing methods are able to achieve non-trivial performance, \shortname forms a challenging open problem. \shortname is available at \benchmarkurl.

%% file: 20-intro.tex
\section{Introduction}\label{sec:introduction}

The ability to reason about natural language has the potential to transform how reinforcement learning (RL) is used to train grounded agents.  
Language provides an expressive and accessible conduit for task specification, enabling agents to address a broad set of tasks, rather than to learn single-behavior policies. 
At the same time, RL is a promising framework to study core grounded language learning problems within interactive environments.

Prerequisite to realizing this potential are expressive benchmark environments, as has been instrumental for progress in RL more broadly. 
However, natural language poses unique challenges to building such benchmarks.
Beyond the design of the environment itself, which must support rich linguistic reasoning, accurate reward computation requires resolving language semantics. 
Existing approaches adopt different strategies to address this issue, most often by using synthetic language~\citep[e.g.,][]{DBLP:conf/ijcai/CoteKYKBFMHAATT18,CoReyes2019:language-meta-rl}, or by removing the language problem from reward computation by restricting to a single goal state~\cite[e.g.,][]{Anderson:18r2r,chen2019touchdown}. 
While these approaches open new research avenues, both have significant drawbacks. 
The simplifications of synthetic language limit the relevance of methods and results to the complexities of human language. 
Single-goal formulations forgo or restrict language's ability to efficiently abstract over many possible solutions, a core argument for its potential to RL.

\begin{figure}[t!]
  \centering
  \includegraphics[width=0.95\linewidth,clip,trim=57 63 57 3]{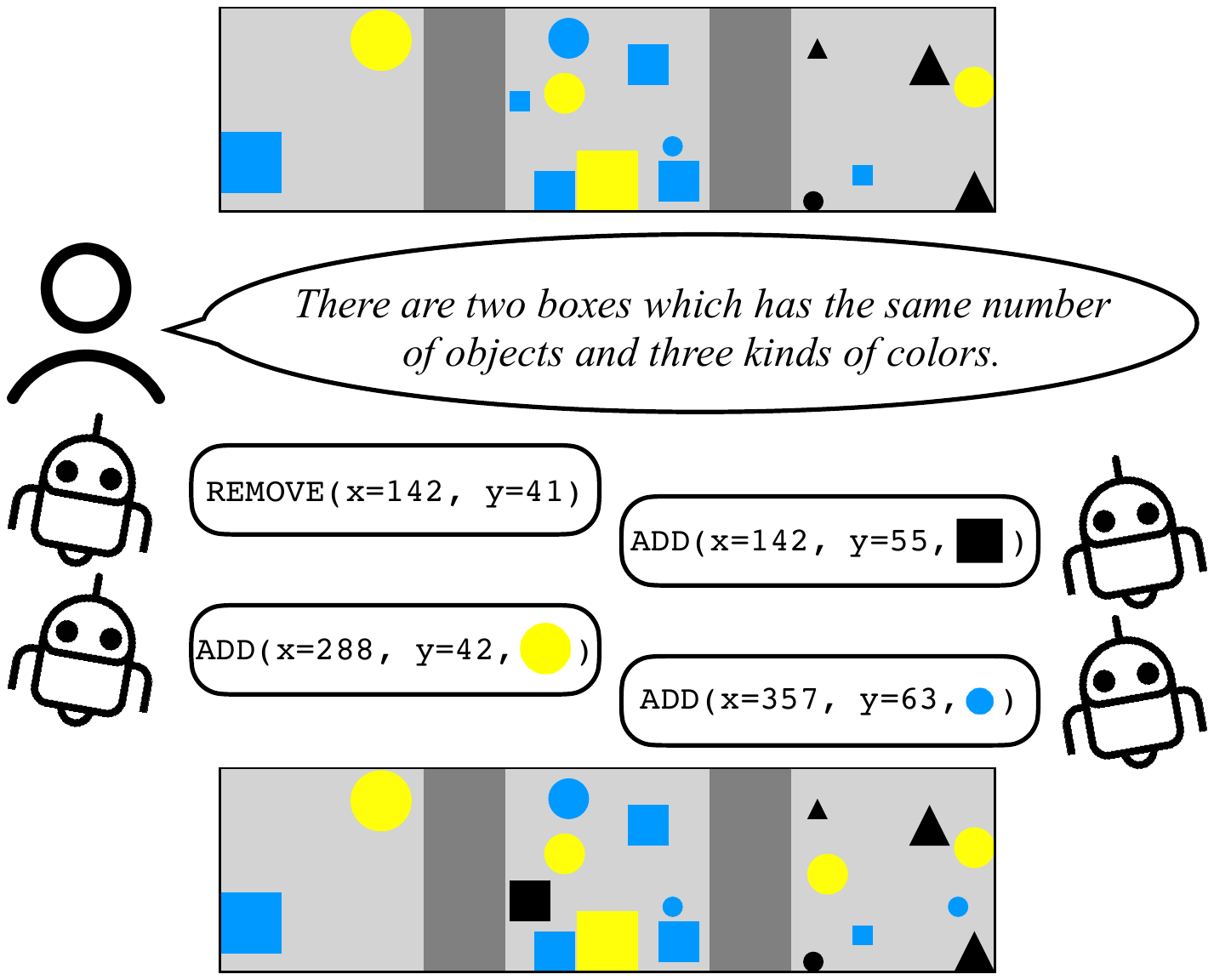}
  \caption{
  An illustrative example of \shortname. Given a natural language statement with a target truth value, \true in this example (omitted from the figure), and an initial image, the agent needs to manipulate objects in the environment so that the truth value of the statement with respect to the image is the target truth value.}\label{fig:main}
  \vspace{-10pt}
\end{figure}

We present \shortname, an RL benchmark where an agent manipulates a visual environment by adding and removing objects conditioned on a natural language statement. 
The agent's goal is to modify the environment so that a given statement will have a pre-specified truth-value with regard to the environment (i.e., the constraints specified in the language are either satisfied or violated). 
\autoref{fig:main} illustrates the scenario. 
\shortname includes highly-compositional and semantically-diverse natural language statements and visual environments from the Natural Language for Visual Reasoning (\nlvr) corpus~\citep{suhr2017corpus}, that combine with a configurable environment backbone to create thousands of Markov Decision Processes (MDP) of varying complexity.

A key challenge in constructing \shortname is accurate reward computation. 
Because of the flexibility of the environments and language, there are many possible equally correct termination states (i.e., goals) for each MDP. 
Correct reward computation at every possible state requires taking into account the semantic nuances of the highly-compositional language. 
We address this by annotating all statements with executable Python programs representing their meaning, effectively creating a supervised semantic parsing corpus~\citep[e.g.,][]{Zelle:96,Zettlemoyer:05}. The executable programs allow for exact and reliable reward computation at every possible state.

Our experiments with \shortname show that existing models demonstrate non-trivial performance given sufficient training time, with multi-modal pre-training providing further benefit. 
However, there remains significant room for improvement. 
For example, on the simplest configuration, our agent can solve 76{.}23\% of the test environments, but performance drops significantly to only 16{.}81\% on the most complex configuration. 
Our experiments also confirm the importance of modeling language meaning for reward computation in learning. 
The \shortname benchmark and trained models are available under the MIT license at \benchmarkurl.

%% file: 25-related.tex
\section{Related Work}\label{sec:related}

RL research has benefited greatly from benchmarks such as Atari~\cite{bellemare2013arcade} and MuJoCo~\cite{todorov2012mujoco}. Although both are synthetic with limited potential to train policies directly applicable to realistic environments, their accessibility and focus on core RL problems have made them impactful drivers of algorithm development. 
For example, the first demonstration of an effective neural network policy~\cite{mnih2013playing} and the development of proximal policy optimization~\cite[\ppo;][]{schulman2017proximal} both used these benchmarks, and both were later shown to generalize to more complex scenarios. 
\shortname is inspired by these benchmarks. 
Rather than aiming for training models that transfer to realistic domains, it aims to enable fundamental algorithmic development by emphasizing semantic diversity and compositionality within an accessible benchmark.

There is significant interest in RL for training policies conditioned on natural language. 
This includes multiple efforts at developing RL environments with language-informed tasks, mainly via grounded language learning in 2D or 3D environments or using text games. 
Oftentimes, using synthetic language~\citep{DBLP:conf/emnlp/NarasimhanKB15,Johnson2017:symbolic-clevr,DBLP:conf/iclr/Chevalier-Boisvert19, DBLP:journals/corr/abs-2004-07200, DBLP:conf/ijcai/CoteKYKBFMHAATT18, DBLP:conf/emnlp/UrbanekFKJHDRKS19,DBLP:journals/corr/HermannHGWFSSCJ17,CoReyes2019:language-meta-rl,DBLP:conf/aaai/HausknechtACY20,jiang2020wordcraft}. 
Although synthetic language allows studying  the problem of learning high-level concepts, many of the complexities of natural language may be stripped away, and such approaches run the risk of reducing the language learning challenge to reverse engineering the hand-crafted generation process. 
In contrast, \shortname is based on semantically-diverse human-written language grounded in a visual environment, and requires both the ability of reasoning over highly-compositional language including sets and spatial relations, and precise alignment between statements and states.

Another approach is to simplify the task so only a few annotated termination states or trajectories are correct~\citep{Anderson:18r2r,chen2019touchdown,Ku2020:room-across-room}. 
This forgoes much of the abstractive potential of natural language, where it can succinctly define extremely large set of states. 
Thereby reducing the utility of language, and not exposing learning algorithms to some of the core challenges it introduces to the RL problem. 
\shortname does not adopt such simplifications. Our experiments show the importance of considering language meaning for reward computation when there is a large set of valid goal states.

Alternatively, other benchmarks are created by first generating the target sequence of decisions (i.e., task demonstration), and then soliciting post-hoc instructional language~\citep{Shridhar2020:alfred,DBLP:conf/iclr/ShridharYCBTH21,hanjie2021grounding}. 
This process uses human-written language, but retains the regularities of the demonstration generation procedure. 
\shortname uses language from \nlvr, which was crowdsourced via a contrastive task that was shown to elicit high semantic diversity.

%% file: 35-background.tex
\section{Background: the \nlvr Corpus}\label{sec:data:nlvr}

\shortname uses data from the \nlvr corpus~\citep{suhr2017corpus}. 
\nlvr was initially created as a supervised learning benchmark. 
We formalize an interactive task using the \nlvr data and collect additional annotations for reward computation.

\nlvr includes human-written natural language statements paired with synthetic images. Each pair is annotated with the boolean truth-value of the statement with regard to the image (i.e., \true if the statement is true with regard to the image, or \false otherwise). 
The images are designed to support complex reasoning, including about spatial and set relations. 
The original learning task posed by \nlvr is to classify statement-image pairs as \true to indicate the statement is true with regard to the image, or \false otherwise. %
\nlvr has been studied extensively~\citep{suhr2017corpus, tan-bansal-2018-object, goldman-etal-2018-weakly, pavez-etal-2018-working, yao-etal-2018-cascaded, DBLP:conf/iclr/HudsonM18, perez2018film, dasigi-etal-2019-iterative, Zheng2020:webly-sup-vis-reasoning,gupta-etal-2021-enforcing}, and a separate version using photos was also released~\citep{suhr-etal-2019-corpus}.\footnote{We do not use the photographic NLVR2 in this work.}

Qualitative analysis of the \nlvr data \citep[Table~2 in][]{suhr2017corpus} showed it to provide diverse representation of semantic and compositional phenomena, including requiring joint visual-linguistic reasoning about spatial relations, quantities, and set relations. 
\nlvr also provides an underlying structured representation for every image, which supports image manipulation. 
The combination of an interface for image manipulation with complex reasoning via natural language makes \nlvr ideal to support an interactive benchmark environment.

%% file: 30-task.tex
\section{The \shortname Benchmark}\label{sec:env_and_tasks}

\input{tikzpictures/41-env_and_tasks}

\shortname consists of a collection of environments that share a common backbone. 
The backbone is a 2D plane that is manipulated by placing and removing objects of different types. 
Each environment instance is a Markov Decision Process (MDP) created by pairing a natural language statement and a target boolean value with a configuration of the shared backbone. 
The goal of the agent is to manipulate the environment by adding and removing objects so that the truth-value of the statement with regard to the environment is the target boolean. 

The learning problem \shortname presents is to induce a policy that generalizes across MDPs. 
We split the MDPs to training, development, and held-out testing sets. 
The training environments are for parameter estimation, while the two other sets are for testing during development and for final held-out testing to report approach performance.\footnote{We recommend reporting both development and held-out test results in future work for easy comparison.}

There are two dimensions of configuration: appearance and starting condition. 
The appearance determines the state space, transition function, and action space. The appearance of the environment can be (a) \envtower: the objects include squares only, and they can be stacked into towers in specific positions only; or (b) \envscatter: objects of different types can be freely distributed.
The two leftmost examples in \autoref{fig:env_task_overview} are from \envtower, and the two rightmost are from \envscatter. 
\envtower gives a more constrained problem with much smaller state and action spaces compared to \envscatter. 

There are two starting conditions, which also determine the agent's goal: (a) \taskscratch: the environment starts without any objects and the goal is to modify it so that the statement's truth-value is \true{}; or (b) \taskflipit: the environment starts with a set of objects and the agent's goal is to flip the truth-value of the statement, by modifying the environment.
The first row of images in \autoref{fig:env_task_overview} shows start states in both conditions. 
\taskscratch generally only requires adding objects, except in cases of correcting for agent's errors, while \taskflipit requires both adding and removing, because there are already objects present. 

The four configurations are \tost, \toft, \scst, and \scft. 
In our experiments (\autoref{sec:exp}), we observe the different configurations provide different levels of difficulty. For example, \envscatter configurations are generally  harder than \envtower, due to the larger state and action spaces.

Formally, each configuration is a Contextual Markov Decision Process~\citep[CMDP;][]{hallak2015contextual}. %
CMDP is an abstraction over a set of MDPs to account for a context that remains constant throughout an interaction with an MDP.  
The context includes the statement and the target boolean the interaction is conditioned on. 
A CMDP is a tuple $(\contextset, \stateset, \actionset, \mappingfunc(c))$, where $\contextset$ is the context space, $\stateset$ the state space, $\actionset$ the action space, and $\mappingfunc$ a function mapping a context $c \in \contextset$ to an MDP $\mappingfunc(c) = (\stateset, \actionset, \transitionfunc, \rewardfunc, \initialstateset)$. Here, $\transitionfunc: \stateset \times \actionset \rightarrow \stateset$ is a transition function, $\rewardfunc: \stateset \times \actionset \rightarrow \real$ a reward function,  and $\initialstateset$ an initial state distribution. This means that a CMDP is a set of MDPs that share the same states and actions. 
The policy learning problem is to estimate parameters $\params$ of a policy $\policy_\params : \stateset \times \contextset \rightarrow \actionset$, which  maps  the current state and the context underlying the MDP to an action. The policy must generalize across different contexts from $\contextset$.
\autoref{fig:env_task_overview} shows example action trajectories in MDPs for each of the four CMDPs. 
\autoref{tab:stat} shows the number of MDPs in each configuration.\footnote{\label{fn:nlvrexp} \nlvr includes 18{,}322 images. This allows further expanding the number of initial states to 92{,}179 initial states through box element permutations. We do not manipulate this property in this work, but future work could take advantage of it. Our reward computation is invariant to such permutations.}

\paragraph{Contexts $\contextset$}
A context $c  \in \contextset$ is a pair $c = (\tokenseq, \targetbool)$, where $\tokenseq$ is a natural language statement and $\targetbool \in \{\true, \false\}$ is a target boolean value for the statement $\tokenseq$ with respect to the state $\state$. 
The set of statements is predefined for \envtower and \envscatter based on the \nlvr data, but identical across the choice of \taskscratch and \taskflipit. 
The target boolean value in \taskscratch is always \true. In \taskflipit, the target boolean value is either \true or \false. 
Depending on the context, different types of reasoning are required. For example, in the second column of \autoref{fig:env_task_overview}, the statement \nlstring{there is no black block as the top of a tower with at most three blocks} requires reasoning about negation, soft cardinality, color, and position, while the statement in the third column \nlstring{there is a box with 2 triangles of same color nearly touching each other} requires a comparison and to reason about several object attributes (shape, color, position). Both require high-level relational reasoning about single objects or sets. 

\input{tables/41-data_stats}

\paragraph{States $\stateset$}
A state $\state \in \stateset$ is an RGB image. 
Images in \shortname are divided into three box regions of identical dimensions by two dark gray separators (\autoref{fig:env_task_overview}). 
The objects in \shortname have three properties, each can take multiple values: shape ($\act{CIRCLE}$, $\act{SQUARE}$, or $\act{TRIANGLE}$), color ($\act{BLACK}$, $\act{BLUE}$, or $\act{YELLOW}$), and size ($\act{SMALL}$, $\act{MEDIUM}$, or $\act{LARGE}$). 
In \envtower, states are constrained to have stacks of up to four $\act{SQUARE}$s of $\act{MEDIUM}$ size and any color at the center of each box. 
\envscatter states support all object shapes, sizes, and colors, and they may be positioned freely.  
In both conditions, objects cannot cross image boundaries or into the separators. 
The choice of starting condition between \taskscratch or \taskflipit does not influence the state space.

\paragraph{Actions $\actionset$ and Transitions $\transitionfunc$}

There are three action types $\act{STOP}$, $\act{ADD}$, and $\act{REMOVE}$.  
$\act{STOP}$ terminates the episode. 
The truth-value of the statement is only evaluated and compared to the target boolean after the $\act{STOP}$ action is taken.
$\act{ADD}$ adds objects to the environment, and $\act{REMOVE}$ removes objects.
$\act{ADD}$ and $\act{REMOVE}$ take arguments that differ between \envtower and \envscatter:

\textbf{\envtower:} 
Both $\act{ADD}$ and $\act{REMOVE}$ take a position argument, which has three possible values corresponding to the three box regions. Objects are added or removed at the top of the stack. Adding an object on top of a stack of four objects or removing an object from an empty box are both invalid actions. $\act{ADD}$ also takes a color argument. For example, the first action on the left trajectory in \autoref{fig:env_task_overview} is adding a yellow square in an empty box. Including $\act{STOP}$, there are $1+(3+1)\times 3 = 13$ actions.

\textbf{\envscatter:} Unlike \envtower, objects of any type can be placed freely in the box regions. Both $\act{ADD}$ and $\act{REMOVE}$ take 2D coordinates that specify pixel location. Adding an object places it so that its top-left coordinates are the given coordinates. Removing an object will remove the object at the given coordinates. Adding also requires specifying the shape, color, and size. The action is invalid if adding results in objects' overlap or boundary crossing with the separators or image boundaries. Removing from a position that does not include an object is also an invalid action. The native resolution of images in \shortname is 380$\times$100 pixels. Including $\act{STOP}$, there are $1 + (380\times100) \times ((3\times3\times3) + 1)  = 1{,}064{,}001$ actions. Because of the extremely large action space, \shortname also supports a simplification through a coarser grid system for \envscatter that is automatically mapped to the original resolution (\aautoref{sec:app:env:grid}). The grid simplification includes heuristics that assist in identifying locations to place objects in the original pixel space or objects to remove once a grid cell is selected. 
In our experiments (\autoref{sec:exp}), we use a grid simplification of 19$\times$5, giving a total of 2{,}661 actions. The difficulty of \envscatter can be adjusted by modifying the grid size, or acting at the original resolution.

The transition function $\transitionfunc: \stateset \times \actionset \rightarrow \stateset$ depends on the choice between \envtower and \envscatter configurations, because this choice determines the action space. Similar to the action space, the transitions in \envtower are more constrained compared to \envscatter. The transition function does not modify the context, which is fixed for a given MDP.

\paragraph{Reward Function $\rewardfunc$} 

The reward function $\rewardfunc$ is computed with respect to the context $\context=(\tokenseq, \targetbool)$, and is based on evaluating the truth-value of the  natural language statement $\tokenseq$ with respect to a state $\state$, and comparing it to the target boolean $\targetbool$. 
\shortname includes an evaluation function $\logicalformfunc : \stateset \times \actionset \rightarrow \{\true, \false\}$ for every statement $\tokenseq$. 
\autoref{sec:data} describes how we create the evaluation functions. 

The agent receives a positive reward for terminating the episode using the $\stopaction$ action if the evaluation $\logicalformfunc(\state)$ is equal to the target boolean $\targetbool$. If $\logicalformfunc(\state)$ does not equal $\targetbool$ when taking the $\stopaction$ action, the agent receives a negative reward. 
If the episode terminates because the current time step $t$ reached the action horizon $\hor$ or because of an invalid action, the agent also receives a negative reward. 
Action validity depends on the current state $\state$ and on the configuration, because $\envtower$ and $\envscatter$ have different action spaces. For example, in $\envtower$, adding an object to a box (e.g., $\act{ADD}(\act{MIDDLE},\act{BLUE})$) is only valid if the box has less than four objects, because towers have a maximum height of four. 
There is also a verbosity penalty of $\delta$. 
Formally, the reward is:

\vspace{-10pt}
\begin{small}
\begin{align}
    \rewardfunc(\state, \action) &= 
        \begin{cases}
            1.0 & \action = \stopaction \wedge \logicalformfunc(\state) = \targetbool \\
            -1.0 & a = \stopaction \wedge \logicalformfunc(\state)\neq \targetbool \\ -1.0 & (a \text{ is invalid in } \state) \lor (t = \hor)\\
            -\delta & \text{otherwise}\ %
        \end{cases}\;\;.
\end{align}
\end{small}

\paragraph{Initial State Distribution $\initialstateset$}

The initial state distribution $\initialstateset$ is parameterized by the context $\context \in \contextset$, and differs between $\taskscratch$ and $\taskflipit$. 
In $\taskscratch$, the agent modifies an empty environment to satisfy the truth-condition of the statement $\tokenseq$ in the context $\context$, so the initial state $\state_0$ is always an empty image. 
The set of initial states $\initialstateset$ for every context $\context \in \contextset$ is the set of images associated with the statement $\tokenseq$ in the NLVR data. This set includes between 1 to 43 images. 
\autoref{tab:stat} shows the total number of initial states in each configuration.

%% file: tikzpictures/41-env_and_tasks.tex
\begin{figure*}[t]
\centering

\begin{tikzpicture}[scale=0.85, transform shape]

\node[align = center, font=\footnotesize\linespread{0.5}, minimum height=3.2em] at (0, 0) (TST0) {One tower has exactly 1 \\ black block and 1 yellow \\ block};
\node[align = center, below=0em of TST0, font=\footnotesize\linespread{0.5}] (TSB0) {Starting label: \false\\Target label: \true};
\node[rectangle, draw, minimum width=7em, minimum height=4.8em, fit=(TST0) (TSB0)] (TSR0) {};
\node[below=0.6em of TSR0] (TS0) {\includegraphics[width=.21\textwidth]{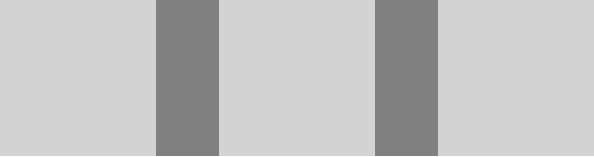}};
\node[left=0em of TS0] (s10) {\footnotesize $s_0$};

\node[right=1.3em of TST0, align = center, font=\footnotesize\linespread{0.5}, minimum height=3.2em] (TFT0) {There is no black block as \\the top of a tower with \\at most three blocks.};
\node[align = center, below=0em of TFT0, font=\footnotesize\linespread{0.5}] (TFB0) {Starting label: \false\\Target label: \true};
\node[rectangle, draw, minimum width=7em, minimum height=4.8em, fit=(TFT0) (TFB0)] (TFR0) {};
\node[below=0.6em of TFR0] (TF0) {\includegraphics[width=.21\textwidth]{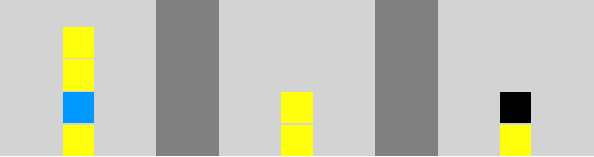}};

\node[right=1.3em of TFT0, align = center, font=\footnotesize\linespread{0.5}, minimum height=3.2em] (SST0) {There is a box with 2 \\triangles of same color \\nearly touching each other.};
\node[align = center, below=0em of SST0, font=\footnotesize\linespread{0.5}] (SSB0) {Starting label: \false\\Target label: \true};
\node[rectangle, draw, minimum width=7em, minimum height=4.8em, fit=(SST0) (SSB0)] (SSR0) {};
\node[below=0.6em of SSR0] (SS0) {\includegraphics[width=.21\textwidth]{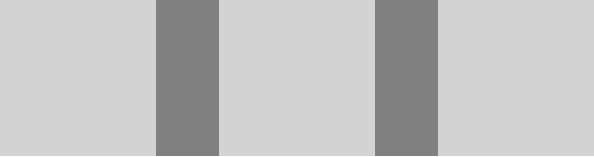}};

\node[right=1.3em of SST0, align = center, font=\footnotesize\linespread{0.5}, minimum height=3.2em] (SFT0) {None of the yellow triangles \\are touching the edge};
\node[align = center, below=0em of SFT0, font=\footnotesize\linespread{0.5}] (SFB0) {Starting label: \true\\Target label: \false};
\node[rectangle, draw, minimum width=7em, minimum height=4.8em, fit=(SFT0) (SFB0)] (SFR0) {};
\node[below=0.6em of SFR0] (SF0) {\includegraphics[width=.21\textwidth]{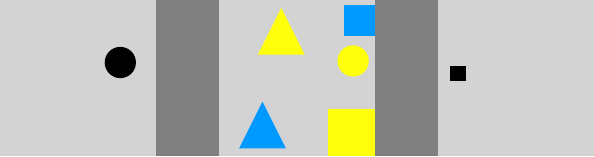}};

\node[left=0em of TSR0] (c10) {\footnotesize $c$};

\node[above=0.5em of TSR0] (l1) {\small\textbf \tost};
\node[above=0.5em of TFR0] (l2) {\small\textbf \toft};
\node[above=0.5em of SSR0] (l3) {\small\textbf \scst};
\node[above=0.5em of SFR0] (l4) {\small\textbf \scft};

\node[below=3.8em of TS0] (TS1)
{\includegraphics[width=0.21\textwidth]{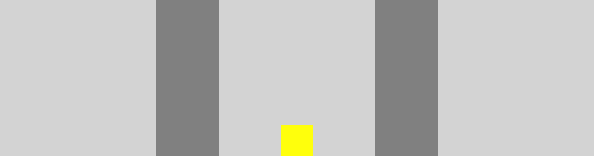}};
\node[left=0em of TS1] (s11) {\footnotesize $s_1$};

\node[below=3.8em of TF0] (TF1)
{\includegraphics[width=0.21\textwidth]{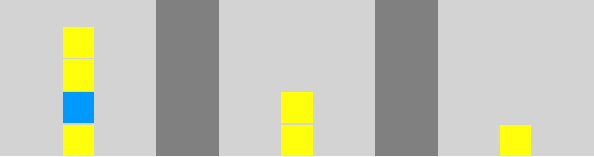}};

\node[below=3.8em of SS0] (SS1)
{\includegraphics[width=0.21\textwidth]{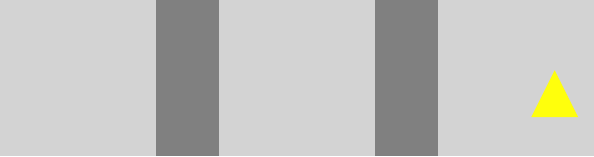}};

\node[below=3.8em of SF0] (SF1)
{\includegraphics[width=0.21\textwidth]{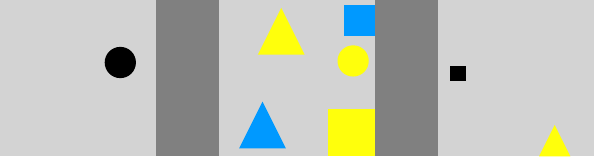}};

\draw[-stealth, thick] (TS0) -- node[left, align=left, font=\footnotesize\linespread{0.5}] {$a_{0} \sim \policy(s_{0}, c)$\\ $r_{0} \sim \rewardfunc(s_{0}, a_{0})$} (TS1);
\draw[-stealth, thick] (TS0) -- node[right, align=left, font=\footnotesize\linespread{0.5}] {$a_{0}$ = $\act{ADD(}$\\$\act{MIDDLE,}$\\$\act{YELLOW)}$} (TS1);

\draw[-stealth, thick] (TF0) -- node[right, align=left, font=\footnotesize\linespread{0.5}] {$a_{0}$ = $\act{REMOVE(}$\\$\act{RIGHT)}$} (TF1);

\draw[-stealth, thick] (SS0) -- node[right, align=left, font=\footnotesize\linespread{0.5}] {$a_{0}$ = $\act{ADD(}$\\$\act{x_{0}, y_{0},}$\\$\act{TRIANGLE,}$ \\ $\act{YELLOW,LARGE)}$} (SS1);

\draw[-stealth, thick] (SF0) -- node[right, align=left, font=\footnotesize\linespread{0.5}] {$a_{0}$ = $\act{ADD(}$\\$\act{x_{0}, y_{0},}$\\$\act{TRIANGLE,}$ \\ $\act{YELLOW,MEDIUM)}$} (SF1);

\node[below=3.8em of TS1] (TS2)
{\includegraphics[width=0.21\textwidth]{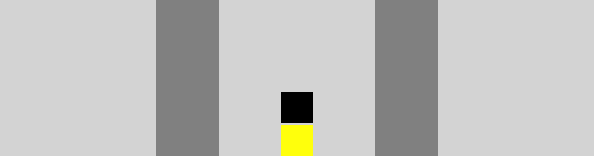}};
\node[left=0em of TS2] (s12) {\footnotesize $s_2$};

\node[below=3.8em of TF1] (TF2)
{\includegraphics[width=0.21\textwidth]{img/tf_1.pdf}};

\node[below=3.8em of SS1] (SS2)
{\includegraphics[width=0.21\textwidth]{img/ss_1.pdf}};

\node[below=3.8em of SF1] (SF2)
{\includegraphics[width=0.21\textwidth]{img/sf_1.pdf}};

\draw[-stealth, thick] (TS1) -- node[left, align=left, font=\footnotesize\linespread{0.5}] {$a_{1} \sim \policy(s_{1}, c)$\\ $r_{1} \sim \rewardfunc(s_{1}, a_{1})$} (TS2);
\draw[-stealth, thick] (TS1) -- node[right, align=left, font=\footnotesize\linespread{0.5}] {$a_{1}$ = $\act{ADD(}$\\$\act{MIDDLE,}$\\$\act{BLACK)}$} (TS2);

\draw[-stealth, thick] (TF1) -- node[right, align=left, font=\footnotesize\linespread{0.5}] {$a_{1}$ = $\stopaction()$} (TF2);

\draw[-stealth, thick] (SS1) -- node[right, align=left, font=\footnotesize\linespread{0.5}] {$a_{1}$ = $\act{ADD(}$\\$\act{x_{1}, y_{1},}$\\$\act{TRIANGLE,}$ \\ $\act{YELLOW,SMALL)}$} (SS2);

\draw[-stealth, thick] (SF1) -- node[right, align=left, font=\footnotesize\linespread{0.5}] {$a_{1}$ = $\stopaction()$} (SF2);

\node[below=3.8em of TS2] (TS3)
{\includegraphics[width=0.21\textwidth]{img/ts_2.pdf}};
\node[left=0em of TS3] (s13) {\footnotesize $s_3$};

\node[below=3.8em of SS2] (SS3)
{\includegraphics[width=0.21\textwidth]{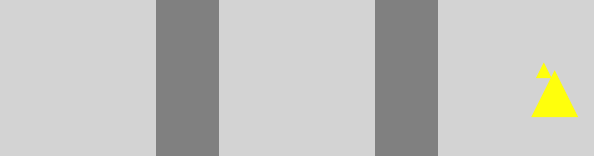}};

\draw[-stealth, thick] (TS2) -- node[left, align=left, font=\footnotesize\linespread{0.5}] {$a_{2} \sim \policy(s_{2}, c)$\\ $r_{2} \sim \rewardfunc(s_{2}, a_{2})$} (TS3);

\draw[-stealth, thick] (TS2) -- node[right, align=left, font=\footnotesize\linespread{0.5}] {$a_{2}$ = $\stopaction()$} (TS3);

\draw[-stealth, thick] (SS2) -- node[right, align=left, font=\footnotesize\linespread{0.5}] {$a_{2}$ = $\stopaction()$} (SS3);

\end{tikzpicture}

\caption{Examples from the four CMDP configurations.
Each example is conditioned on a context $\context=(\tokenseq, \targetbool)$, and starts with a state $\state_0$, sampled from the initial state distribution $\initialstateset$. 
For example, for \tost (left column), the context $\context$ pairs the statement \nlstring{one tower has exactly 1 black block and 1 yellow block} with the target boolean \true. The initial state $\state_{0}$ is an image with three empty grey box regions separated by darker grey separators.
The agent $\policy$ is given $(\state_0, c)$, and samples an action $\action_0 \sim \policy(\state_0, c)$. The environment transitions to the next state $s_1$, while the context remains the same.
This process continues until the agent selects the $\stopaction$ action.}\label{fig:env_task_overview}
\vspace{-10pt}
\end{figure*}

%% file: tables/41-data_stats.tex
\begin{table*}[t]
\centering
\footnotesize
\setlength{\tabcolsep}{5pt}
\begin{tabular}{lccccccccc}
\toprule
   \multicolumn{1}{c}{} & \multicolumn{1}{c}{\textbf{\tost}}  && \multicolumn{2}{c}{\textbf{\toft}}            && \multicolumn{1}{c}{\textbf{\scst}}  && \multicolumn{2}{c}{\textbf{\scft}}            \\ 
  \cmidrule{2-2} \cmidrule{4-5} \cmidrule{7-7} \cmidrule{9-10}  
    & MDPs && MDPs & Init.  && MDPs && MDPs & Init.  \\ 
    \midrule
Train & 989 &&  1{,}910   & 5{,}704 && 1{,}241    && 2{,}340   & 6{,}696  \\ %

Dev &  163      && 317      & 676  && 87       &&  164  & 313  \\ %

Test & 324      && 619    & 1{,}383  && 155      && 285      & 591  \\ \midrule

Total & 1{,}476    && 2{,}846    & 7{,}763 && 1{,}483    && 2{,}789   & 7{,}600 \\ %
\bottomrule
\end{tabular}
\caption[Caption for stats table]{Data statistics per CMDP configuration and data split. The number of MDPs corresponds to the number of contexts under each CMDP. For \taskflipit, ``Init.'' corresponds to the total number of initial states across all MDPs for this CMDP.$^{\ref{fn:nlvrexp}}$ The number of initial states and MDPs is equal for \taskscratch CMDPs.}\label{tab:stat}
\vspace{-10pt}
\end{table*}

%% file: 40-dataset.tex
\section{The \shortname Data and Annotation}\label{sec:data}

We use the \nlvr data to create each of the CMDPs (\autoref{tab:stat}). 
\taskscratch CMDPs include contexts for all natural language statements from \nlvr, each paired with the empty initial state containing no shapes (\autoref{fig:env_task_overview}, left and center-right columns).
\taskflipit CMDPs include the natural language statements with their corresponding images, both from \nlvr. The images are used as initial states. The target boolean is set so that the initial state does not fulfil it.
The split between \envtower and \envscatter also follows from \nlvr. Statements corresponding to \envtower images in \nlvr are included in our \envtower CMDPs, and the same for \envscatter sentences.

\nlvr has four splits for training, development, public testing, and hidden testing.
We adopt the original training and development sets splits. 
Following the recent public release of the hidden testing set, we merge the public and hidden testing sets into a single public test split.

The \nlvr annotations include the truth-value of each statement with regard to the images paired with it in the data. 
Once we manipulate an image (i.e., change the state in our environment), the truth-value annotation does not necessarily hold. 
A key challenge for creating an interactive environment using this data is an accurate evaluation of the natural language statement for \textit{every} possible state (i.e., image) for reward computation (\autoref{sec:env_and_tasks}). 
We address this by annotating each statement $\tokenseq$ with an executable boolean Python program representing its meaning, $\logicalformfunc$ in \autoref{sec:env_and_tasks}. 
This process is inspired by data annotation for supervised semantic parsing~\cite{Zelle:96,Zettlemoyer:05,Suhr:18context}, where sentences are annotated with formal meaning representations. 

The Python programs operate on the underlying structured representation. 
Each program returns \true for every image that satisfies the constraints specified in the corresponding statement, and \false otherwise. In general, there are many states that satisfy any given statement, many more than provided with the original \nlvr images. 

The programs are written using an API defined over the structured representations. 
We base the API design on the ontology designed for \nlvr's structured representations by \citet{goldman-etal-2018-weakly}, which we extend to include 66 functions. 
\autoref{fig:logical_forms_side} in \autoref{sec:app:data} shows two example programs with their corresponding statements.

We use the freelancing platform Upwork\footnote{\url{https://www.upwork.com}} for annotation. We recruit three programmers based on preliminary screening of their fluency in English and competency in Python. 
We de-duplicate the naturally occurring sentences in the data, and distribute sentences to annotators randomly, each with a single example \nlvr image. 
Each program is evaluated against a corresponding hidden validation set made of all remaining \nlvr images paired with the sentence, and must pass all the tests.
\aautoref{sec:app:data} provides a screenshot of the interface and more details. 
We collect 2{,}666 annotations at a total cost of \$3{,}756, and keep 2{,}661 valid annotations.

%% file: 50-exp.tex
\section{Experiments}\label{sec:exp}

\subsection{Methods}\label{sec:exp:methods}

We experiment with each of the four CMDPs separately, training on the training split and testing on the development and test splits. 
We sample a validation set from the training split for model selection. 
For \envscatter we use a simplified grid action space of 19$\times$5 (\autoref{sec:env_and_tasks}). Each grid cell is 20$\times$20 pixels. We set the action horizon $\hor=12$. 
\aautoref{sec:app:exp} provides implementation details.

We use \ppo~\citep{schulman2017proximal} for parameter estimation,\footnote{We use the \ppo implementation of \cite{kostrikov2018pytorch}.} with a separate network as a critic. The critic network is identical to the policy, except that we add a $\tanh$ activation for the value output. 
Because of the large action space, especially for \envscatter, the agent rarely observes positive reward, which requires taking a $\stopaction$ action at an appropriate state. 
We design a simple variant of \ppo called \ppowsf (\ppo with stop forcing) to study this issue. 
\ppowsf is identical to \ppo, except that during training, we mask all actions except $\stopaction$ when the agent reaches a state where selecting $\stopaction$ will give a positive reward. 
This modification is present only during training. All testing is done under the same conditions, without stop forcing. 

We also study the importance of our annotation for reward computation using a reward function for \taskscratch that does not require any annotation beyond what is already in \nlvr. The reward uses \nlvr images that are associated with each statement (between 1--43) and are labeled with the target boolean.\footnote{We dropped 1\% of MDPs without a single target state.} Instead of testing the state using a program, it compares the state to the available \nlvr images, and only if it equals one of them, the learner receives a positive task completion reward. 

We experiment with three models: \cnnbase, \cnnbaseten and \vilt.\footnote{\autoref{sec:app:other_models} describes preliminary experiments with one more model.}
In \cnnbase and \cnnbaseten, we process the statement $\tokenseq$ using BERT~\citep{devlin-etal-2019-bert}, and do mean pooling across all layers and tokens to get the statement representation. 
We use a three-layer CNN~\citep{fukushima1982neocognitron} in \cnnbase to embed the image of the current state $\state$, and a ten-layer CNN in \cnnbaseten. 
We concatenate the statement and image representations with  an embedding for the target boolean $\targetbool$, and use a multi-layer perceptron to compute the action distribution. 
\vilt is a pretrained multi-modal Transformer that jointly processes text and image inputs~\citep{pmlr-v139-kim21k}. 
We create a sequence of tokens by concatenating the statement, a token for the target boolean, and image patches, separated by special tokens. The image patches are the same size as the 19$\times$5 grid cells, including in \envtower, where the action space does not use a grid.

\input{60-exp-results.tex}

%% file: 60-exp-results.tex
\input{tables/61-table_results}

\subsection{Results and Analysis}\label{sec:exp:results}

\begin{figure}[t]   
  \centering
  \includegraphics[width=1\linewidth]{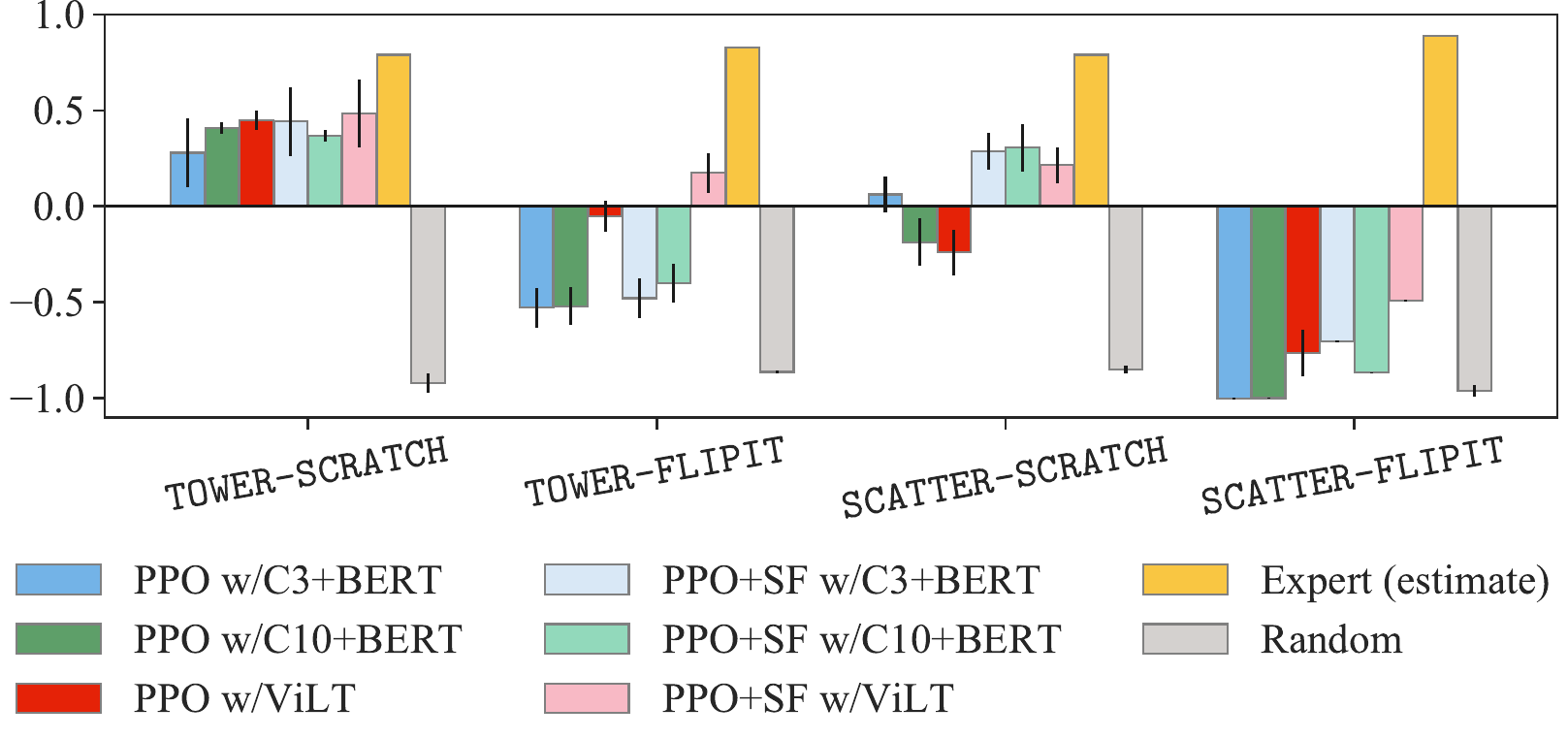}
  \caption{Mean development set rewards, averaged over three runs.}\label{fig:efficiency}
  \vspace{-10pt}
\end{figure}

\begin{figure}[t]
  \centering
  \includegraphics[width=1\linewidth]{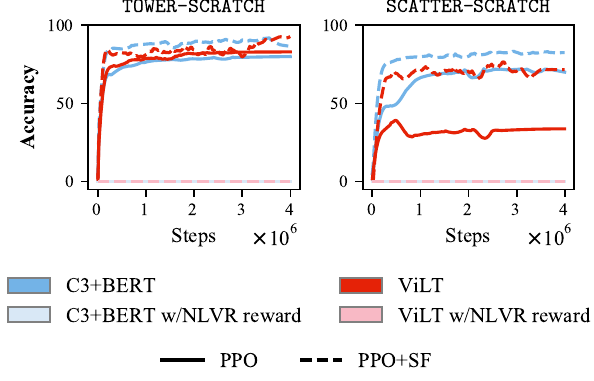}
  \caption{Training curves for \taskscratch, averaged over three runs. All \nlvr reward curves are superposed because accuracy remains zero throughout training.\textsuperscript{\ref{fn:nopatience}}}
  \label{fig:training_curves_with_nlvr_reward}
\end{figure}

\begin{figure}[t]
  \centering
  \includegraphics[width=.85\linewidth]{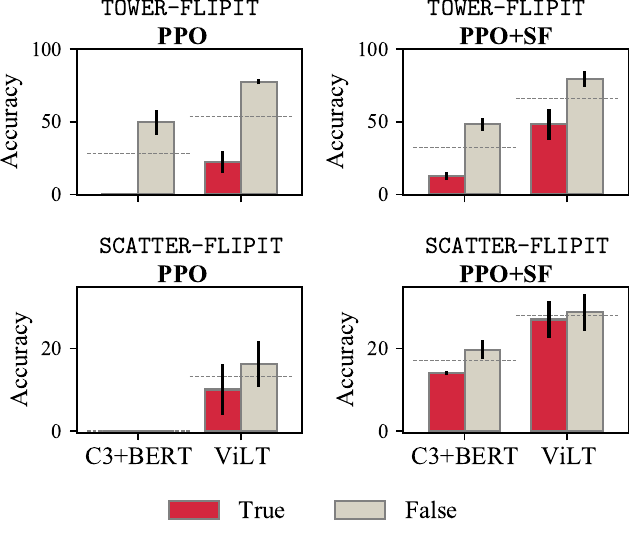}
  \caption{Mean development set accuracies for \taskflipit CMDPs, averaged over three runs, reported according to the value of the context target boolean (\textbf{Red} for \true, \textbf{Gray} for  \false). \textbf{Dashed gray line}: full development set accuracies.
  }\label{fig:res_flipit}
\end{figure}

\begin{figure}[t!]
  \includegraphics[width=1\linewidth]{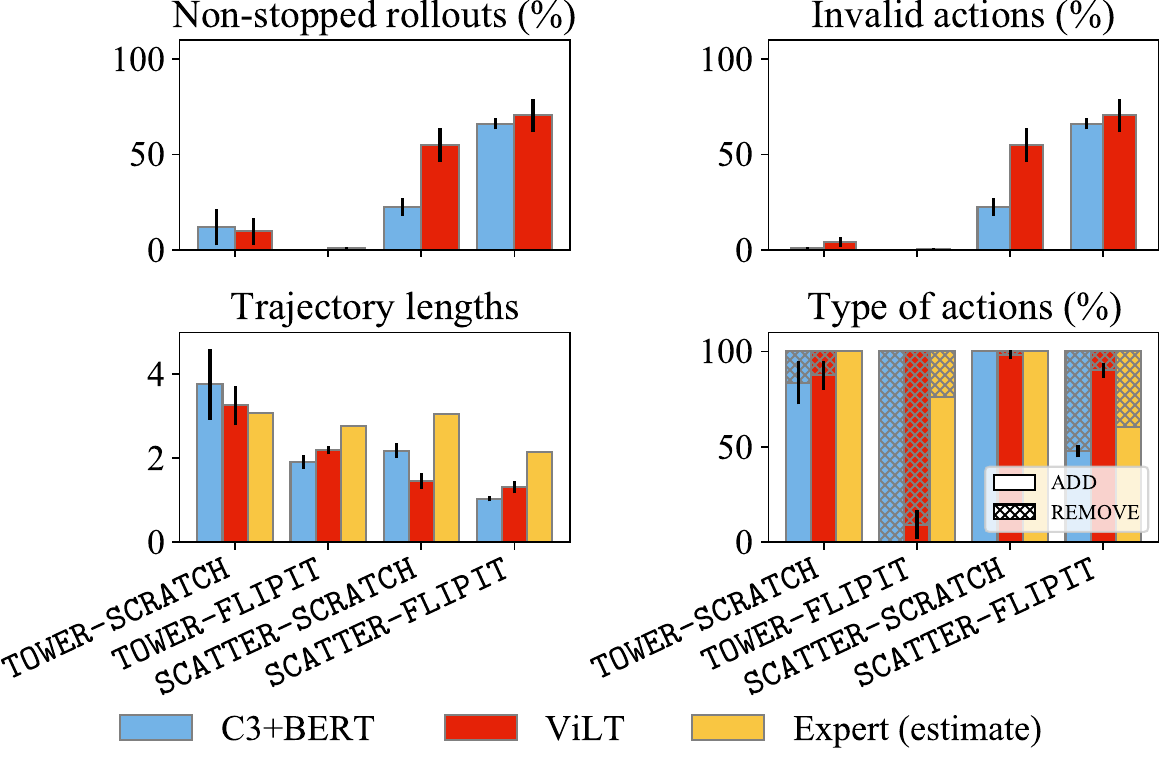}  
  \caption{Mean development statistics for \ppo, averaged over three runs. Clockwise from top left: rollouts without a $\stopaction$ action; rollouts with an invalid action; mean actions per rollout; relative rate of $\act{ADD}$/$\act{REMOVE}$ actions.}\label{fig:analysis_quant}
  \vspace{-10pt}
\end{figure}

\autoref{tab:results} shows task-completion accuracies for all CMDPs, and \autoref{fig:efficiency} shows reward statistics. 
We observe only minor differences between \cnnbase and \cnnbaseten, so conduct the bulk of our analysis on \cnnbase and \vilt. 
\autoref{fig:training_curves_with_nlvr_reward} plots training curves for \taskscratch.\footnote{We plot the training curves with no regard to the patience stopping criteria we use for model selection. This better reflects long term trends.\label{fn:nopatience}} 
\autoref{fig:res_flipit} breaks down development set accuracies for \taskflipit CMDPs by the target boolean, and \autoref{fig:analysis_quant} shows development rollout statistics for \ppo.\footnote{\aautoref{sec:app:results} provides further analysis, including error analysis, performance breakdown by semantic phenomena, and rollout statistics for \ppowsf.} 
We sample 50 development examples for each CMDP, and annotate them with expert\footnote{The expert is an author of this paper.} trajectories to estimate the expert reward and rollout statistics. All expert rollouts are successful. 

Overall, we observe stronger task-completion performance (\autoref{tab:results}) on \envtower CMDPs compared to \envscatter, especially with \vilt, which shows stronger performance than \cnnbase and \cnnbaseten in most cases. 
The development rewards (\autoref{fig:efficiency}) and training curves (\autoref{fig:training_curves_with_nlvr_reward}) show similar trends . 
The training curve comparison to the alternative reward that uses \nlvr images instead of our program annotations shows no effective learning. This illustrates the importance of exact reward computation, such as possible with our program annotations. 
The comparison to the estimate of expert rewards shows there remains significant room for improvement across the board. 
Even when the learned policies are able to complete the task, they are doing it inefficiently, so much so that the mean rewards are often negative. 
The mean rewards of the random baseline policy illustrate the task is far from trivial. 
Both task accuracies and reward statistics indicate \tost is the easiest of the CMPDs, and \scft is the hardest. 

The additional guidance of \ppowsf compared to \ppo helps with exploration, especially on \envscatter CMDPs. 
On \scft, \ppowsf improves performance by 13{.}25\% compared to \ppo. 
This illustrates the exploration challenges \envscatter CMDPs pose.

\vilt generally outperforms \cnnbase and \cnnbaseten, except on \scst, where the \vilt policy more often selects invalid actions. 
\vilt general advantage is expected given the joint reasoning architecture and multi-modal pre-training of \vilt. 
\taskflipit policies generally do better on examples with a \false target boolean, except when learning fails (\autoref{fig:res_flipit}). 
The other direction is harder, because the set of states that invalidates a statement is usually larger than the set that validates it, and it generally requires fewer actions to invalidate a statement. 

We observe more rollouts that are terminated either by reaching the action horizon $\hor$ or by taking an invalid action (i.e., without $\stopaction$) on \envscatter CMDPs compared to \envtower (\autoref{fig:analysis_quant}, upper left). 
This difference is partially explained by a higher rate of invalid actions in \envscatter (\autoref{fig:analysis_quant}, upper right), which cause immediate rollout termination. 
On \envscatter CMDPs, where we have a higher rate of invalid actions, the type of invalid actions we most often observe for both models are actions hitting one of the separators, except when learning fails completely.
This is related to the action selection bias of the models, which tend to select some coordinates more often than others. \autoref{sec:app:results:errors} provides further error analysis for both models for \envscatter CMDPs trained with \ppo, and \autoref{sec:app:results:actbias} illustrates the action selection bias.

There is no consistent difference in the length of rollouts between the two models (\autoref{fig:analysis_quant}, bottom left). 
Expert trajectory length is similar on \envtower, where models perform fairly well. However, on \envscatter, where our models are weaker, expert trajectories are significantly longer. This is partially explained by the models not learning to effectively avoid invalid actions, which terminate the execution immediately.  
Using $\act{REMOVE}$ actions is generally difficult for the learned policies. 
\toft is an exception with $\act{REMOVE}$ dominating the rollouts (\autoref{fig:analysis_quant}, bottom right), potentially because removing objects generally provides a more efficient path to flip the boolean value. 
While \ppo policies generate $\act{REMOVE}$ actions for \scft, the extremely low performance indicates that these actions are not used effectively.
Expert statistics indicate that $\act{REMOVE}$ actions are beneficial for \taskflipit CMDPs.

\input{tables/71-ling-analysis_ppo.tex}

We also performed semantic and syntactic analyses using the 200 development examples manually annotated by \citet{suhr2017corpus}. 
\autoref{tab:ling_ppo_short} shows the performance on this data of policies trained with \ppo.
We only include categories with more than 10 instances across all CMDPs. 
\aautoref{sec:app:results:semantics} provides the complete tables with examples, including for \ppowsf. 
The two models mostly follow similar trends with respect to the categories on which they perform above and below overall performance. 
Both models perform better than they do overall on hard cardinality (e.g., \nlstring{\dots exactly four objects \dots}) for \envtower CMDPs, and on presupposition for \taskscratch CMDPs.
However, on spatial relations, both models perform below overall performance for all CMDPs except \tost.

%% file: tables/61-table_results.tex
\begin{table*}[t]
\centering
\footnotesize
\setlength{\tabcolsep}{2pt}
    \begin{tabular}{llccccccccccc}
    \toprule
    && \multicolumn{2}{c}{\textbf{\tost}} && \multicolumn{2}{c}{\textbf{\toft}} && \multicolumn{2}{c}{\textbf{\scst}} && \multicolumn{2}{c}{\textbf{\scft}}
    \\
    \cmidrule{3-4} \cmidrule{6-7} \cmidrule{9-10} \cmidrule{12-13}
    & & Dev & Test && Dev  & Test && Dev & Test && Dev & Test 
    \\ 
    \midrule
     \multirow{2}{*}{\ppo} & \cnnbase & 72{.}80\pmstring{9{.}92}  & 68{.}72\pmstring{6{.}60} && 28{.}11\pmstring{6{.}03}  & 27{.}84\pmstring{4{.}50} && \textbf{59{.}00\pmstring{5{.}42}} & \textbf{68{.}17\pmstring{2{.}89}} && \phantom{0}0{.}00\pmstring{0{.}00}  & \phantom{0}0{.}06\pmstring{0{.}10}\\

    & \cnnbaseten & 80{.}16\pmstring{1{.}77} & 73{.}66\pmstring{3{.}04} && 30{.}13\pmstring{8{.}22} & 28{.}75\pmstring{7{.}42} && 43{.}68\pmstring{7{.}18} & 50{.}97\pmstring{12{.}95} && \phantom{0}0{.}00\pmstring{0{.}00} & \phantom{0}0{.}00\pmstring{0{.}00}\\

    & \vilt & \textbf{81{.}19\pmstring{2.90}}  & \textbf{76{.}23}\pmstring{5.58} && \textbf{53{.}25\pmstring{4.28}}  & \textbf{55{.}19\pmstring{4.92}} && 40{.}23\pmstring{6.99} & 47{.}74\pmstring{11.47} && \textbf{13{.}31\pmstring{6{.}88}}  & \textbf{16{.}81\pmstring{7{.}43}}\\

    \midrule
    
    \multirow{2}{*}{\ppowsf} & \cnnbase & 81{.}80\pmstring{0.94}  & 76{.}54\pmstring{1.85} && 32{.}59\pmstring{3{.}14}  & 29{.}55\pmstring{4{.}74} && 72{.}03\pmstring{3{.}32} & 74{.}41\pmstring{3{.}25} && 17{.}04\pmstring{1{.}82}  & 18{.}16\pmstring{3{.}07} \\

    & \cnnbaseten & 77{.}91\pmstring{2{.}21} & 75{.}62\pmstring{1{.}41} && 37{.}38\pmstring{2{.}02} & 35{.}70\pmstring{2{.}08} && \textbf{73{.}18\pmstring{2{.}65}} & \textbf{77{.}85\pmstring{1{.}}62} && \phantom{0}8{.}52\pmstring{2{.}72} & 10{.}55\pmstring{2{.}45}\\
    
    & \vilt & \textbf{84{.}05\pmstring{3{.}25}}  & \textbf{81{.}48\pmstring{1{.}93}} && \textbf{65{.}68\pmstring{9{.}17}}  & \textbf{65{.}51\pmstring{8{.}43}} && 67{.}43\pmstring{1{.}35} & 73{.}98\pmstring{0{.}30} && \textbf{28{.}01\pmstring{5{.}32}}  & \textbf{30{.}06\pmstring{4{.}51}} \\

    \bottomrule
    \end{tabular}
    \caption{Mean task-completion accuracy and standard deviation computed over three runs for all four CMDPs.}\label{tab:results}
    \vspace{-10pt}
\end{table*}

%% file: tables/71-ling-analysis_ppo.tex
\begin{table*}[t]

\centering
\footnotesize
\setlength{\tabcolsep}{2.3pt}
    \begin{tabular}{lccccccccccc}
    \toprule
    & \multicolumn{2}{c}{\textbf{\tost}} && \multicolumn{2}{c}{\textbf{\toft}} && \multicolumn{2}{c}{\textbf{\scst}} && \multicolumn{2}{c}{\textbf{\scft}}
    \\
    \cmidrule{2-3} \cmidrule{5-6} \cmidrule{8-9} \cmidrule{11-12}
    & Total & Correct \% && Total  & Correct \% && Total & Correct \% && Total & Correct \% 
    \\
    \midrule
    Cardinality (hard) &98 & \textbf{76{.}5}~{\color{lightgray}\vrule}~\textbf{83{.}7} && 480 & \textbf{28{.}9}~{\color{lightgray}\vrule}~\textbf{56{.}2} && 35 & 49{.}5~{\color{lightgray}\vrule}~40{.}0 && 119 & \phantom{0}0{.}0~{\color{lightgray}\vrule}~\textbf{13{.}7}
    \\
    Cardinality (soft) &21 & 68{.}2~{\color{lightgray}\vrule}~81{.}0 && 82 & \textbf{30{.}1}~{\color{lightgray}\vrule}~\textbf{56{.}5} && 11 & \textbf{60{.}6}~{\color{lightgray}\vrule}~24{.}3 && 42 & \phantom{0}0{.}0~{\color{lightgray}\vrule}~10{.}3
    \\
    Existential       &122 & \textbf{75{.}1}~{\color{lightgray}\vrule}~\textbf{81{.}7} && 577 & \textbf{29{.}0}~{\color{lightgray}\vrule}~55{.}1 && 55 & \textbf{55{.}7}~{\color{lightgray}\vrule}~35{.}1 && 192 & \phantom{0}0{.}0~{\color{lightgray}\vrule}~12{.}3
    \\
    Coordination     &19 & \textbf{85{.}9}~{\color{lightgray}\vrule}~\textbf{84{.}2} && 86 & 27{.}1~{\color{lightgray}\vrule}~52{.}7 && 15 & \textbf{64{.}5}~{\color{lightgray}\vrule}~40{.}0 && 55 & \phantom{0}0{.}0~{\color{lightgray}\vrule}~\phantom{0}9{.}7
    \\
    Spatial Relations &94 & \textbf{74{.}8}~{\color{lightgray}\vrule}~\textbf{81{.}6} && 438 & 26{.}6~{\color{lightgray}\vrule}~53{.}1 && 39 & 53{.}8~{\color{lightgray}\vrule}~32{.}5 && 128 & \phantom{0}0{.}0~{\color{lightgray}\vrule}~10{.}1
    \\
    Presupposition    &17 & \textbf{74{.}5}~{\color{lightgray}\vrule}~\textbf{90{.}2} && 74 & 27{.}0~{\color{lightgray}\vrule}~\textbf{54{.}1} && 22 & \textbf{66{.}7}~{\color{lightgray}\vrule}~\textbf{51{.}5} && 78 & \phantom{0}0{.}0~{\color{lightgray}\vrule}~12{.}4
    \\
    \midrule
    \multicolumn{2}{l}{\textbf{Overall}} & 72{.}80~{\color{lightgray}\vrule}~81{.}19 && & 28{.}11~{\color{lightgray}\vrule}~53{.}25 &&  & 59{.}00~{\color{lightgray}\vrule}~40{.}23 &&  & \phantom{0}0{.}00~{\color{lightgray}\vrule}~13{.}31 \\
    \bottomrule
    \end{tabular}
\caption{Performance on a set of development examples annotated for semantic categories by \citet{suhr2017corpus} for both models (\cnnbase~|~\vilt) when trained with \ppo. 
Developement performance refers to mean performance on the respective full development set. Results outperforming dev performance are in bold.}\label{tab:ling_ppo_short}
\end{table*}

%% file: 80-conclusion.tex
\section{Conclusion}\label{sec:conclusion}

We introduce \shortname, an RL benchmark that focuses on natural language visual reasoning. 
\shortname is designed to be accessible for researchers, while still displaying the reasoning richness of natural language. 
It is relatively easy to deploy using the standard Gymnasium API~\citep{Brockman2016:openai-gym}, and has light compute requirements. 
Our data annotation approach allows including expressive and diverse natural language, while still providing accurate and automatic reward computation. 
It also exposes the potential connection between semantic parsing and reward evaluation in RL, thereby outlining how strong semantic parsers can benefit RL benchmarking. 
Our strong baselines illustrate the range of challenges \shortname presents, showing that existing methods can achieve non-trivial performance, but that there remain significant progress to be made. 
Our analysis lays out the framework for studying and reporting these future results. 

\shortname has significant potential beyond the tasks we study. 
It can be used without the language, to create thousands of micro RL tasks requiring set and relational visual reasoning. 
Our annotations form a new semantic parsing corpus with annotated executable meaning representations. The semantic diversity of the data, its executability, and the focus on visual reasoning make it a unique asset in the landscape of corpora for semantic parsing. 
\shortname is also promising for program synthesis guided by natural language~\citep{Wong2021:language-prog-synthesis}.

%% file: 85-limitations.tex
\section{Limitations}\label{sec:limitations}

\shortname uses synthetic visual stimuli, which does not reflect the complexity or characteristics of realistic visual observations. 
This is critical for our ability to control the environment and provide a lightweight and accessible RL benchmark. 
Our goal is not to provide a resource for the development of methods that aim to handle realistic visual input, and \shortname is not suitable for this purpose. 
The limited number of colors, shapes, and sizes used limits the visual and lexical complexity of the data.
The synthetic nature of the data and the modular library of functions we use allow to relatively easily extend the environment (e.g., with new colors). This will require collecting additional natural language data. In this work, we opted to rely on the \nlvr data without further expanding it. 
Some annotators of the original \nlvr data adopted annotation strategies that led to repetition of some common phrases (e.g., starting statements with \nlstring{there is}). 
While this creates some implicit patterns in the data, \citet{suhr2017corpus} showed that \nlvr demonstrates high semantic diversity and compositionality. 
Finally, \shortname includes English data only. Expanding this data to other language is an important direction for future work. 
Translating the data is a feasible low-cost solution, because the program annotations will not require updating.

%% file: 87-ethics.tex
\section*{Ethics Statement}

We paid U.S. standard market wage to our programmers (\autoref{sec:app:data}). The rate was determined by the workers. 
The \shortname environment and data as is are intended to be used for research, including algorithm development and evaluation, and not for development of models to be deployed.

%% file: 90-acks.tex
\section*{Acknowledgements}

This research was supported by ARO W911NF21-1-0106, NSF under grant No. 1750499, and a gift from Open Philanthropy. 
KB is supported by NSF under grant No. 2127309 to the Computing Research Association for the CIFellows Project. 
Results presented in this paper were obtained using CloudBank~\citep{Norman2021:cloudbank}, which is supported by the National Science Foundation under award No. 1925001. 
We thank Alane Suhr, Ge Gao, Justin Chiu, Woojeong Kim, Jack Morris, Jacob Sharf and the Cornell NLP Group for support, comments, and helpful discussions.

%% file: 100-app_task.tex
\section{{\tt SCATTER} Grid Simplification}\label{sec:app:env:grid}

To reduce the large action space of \envscatter,  \shortname allows to simplify the pixel-based action space with a grid that is coarser than the image resolution of 380$\times$100. 
The actions applied in the environment remain in the original resolution, and the translation between the grid system to pixels is done heuristically. Without the heuristics, the transition to a grid coarser than the original image resolution would render many of the MPDs unsolvable.

The heuristics simplify two translation problems: in what pixel exactly to place an object and which object to remove from a grid cell. 
Depending on the grid size, it is possible to add multiple objects in a cell. To find the exact pixel within a cell to add an object, we search for a pixel in the grid box where we can add the object starting from the upper left corner. We can add an object in a pixel if the object fits there without overlapping with other objects, the image boundaries, or the columns. We  also snap objects to touch each other if the distance between them is below a threshold. This is to allow adding objects that \textit{touch} each other, a common constraint in \shortname statements. 
When removing an object from a grid cell, we remove the object with largest overlap  with the cell.

%% file: 110-app_dataset.tex
\section{Natural Language Annotation Details}\label{sec:app:data}

We annotate each natural language statement in the \nlvr corpus with a Python program representing its meaning. The programs return a boolean value, and are executable given the structured representation underlying each image. 
\autoref{fig:logical_forms_side} shows two examples of text statements with their annotated Python programs.

\input{tikzpictures/43-logical_forms_side.tex}

\begin{figure*}
    \centering
    \frame{\includegraphics[width=1\linewidth]{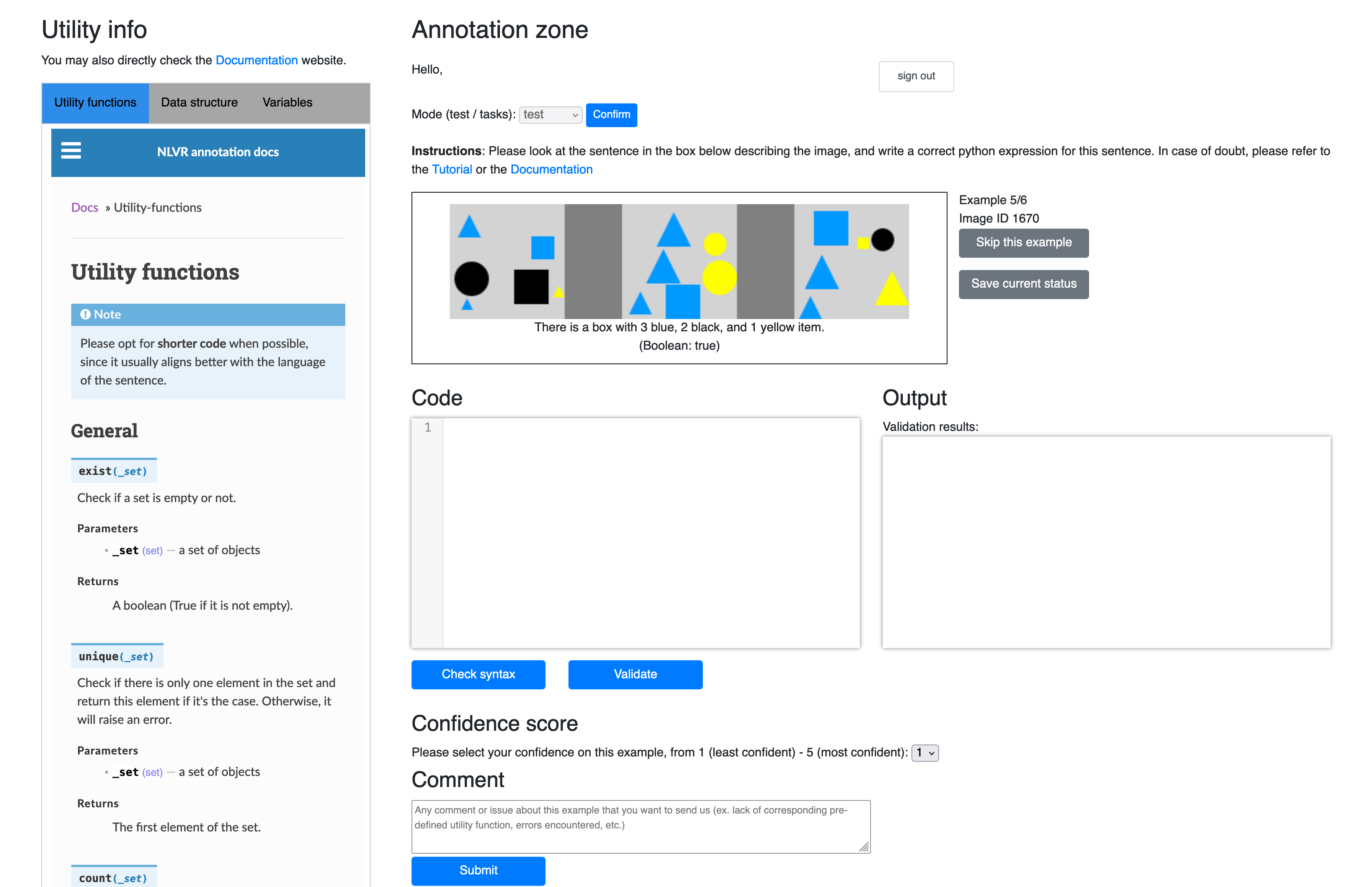}}
    \caption{The annotation interface for collecting the Python program annotations in \shortname.}\label{fig:annotation_env}
\end{figure*}

We provide the annotators with a web-based annotation interface (\autoref{fig:annotation_env}), a tutorial, and an application programming interface (API) presenting a set of functions, classes and objects that they can use for annotation. 
We ask the annotators to prioritize the faithfulness of the program to the natural language sentence and to prefer shorter annotations. 
We also provide them with examples of spurious logical forms and ask them to avoid such expressions. 
Annotators can raise questions.

\autoref{fig:annotation_env} shows the annotation interface for a single sentence. 
For every sentence, annotators are provided with a single example image from \nlvr and an associated boolean value.
Other images for the same statement from \nlvr are used as hidden validation examples. The annotator never sees these images. 

The annotator can validate the program syntax and validate it within the browser. 
The validation executes the program against the given image and all hidden images. 
Validation passes only once the program returns the expected boolean value for all examples, including the visible and the hidden ones. 
The annotator can only submit their annotation after passing the syntax check and validation. 
They can assign a confidence score to their annotation and provide a comment. 

Annotators can skip examples in case of doubt. When skipping, they need to explicitly provide the reason. We assess the annotations by batch, then randomly redistribute the skipped examples or examples with problematic annotations to the annotators after the questions have been solved. 
We iteratively communicate with the workers throughout the entire annotation process.

The annotation was done by four workers, one each from Croatia, India, Ukraine and United States. 
The hourly rate was roughly \$23{.}25 per hour. 
We communicated to the workers the purpose of the data collection and how data will be used at recruiting time.

%% file: tikzpictures/43-logical_forms_side.tex
\begin{figure}

\begin{minipage}{0.4\textwidth}
\begin{tikzpicture}

\node[] at (0,0) (TS0)
{\includegraphics[width=1\textwidth]{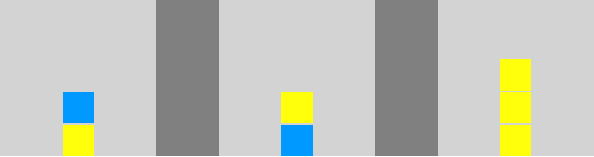}};
\node[align = center, below=-0.3em of TS0, font=\footnotesize\linespread{0.5}] (TST0) {There are two towers with the same height \\ but their base is not the same in color.};
\node[rectangle, draw, minimum width=20em, minimum height=12em, fit=(TS0) (TST0)] (TSR0) {};
\end{tikzpicture}
\end{minipage}
\hfill
\begin{minipage}{0.57\textwidth}\raggedleft
\begin{lstlisting}[language=Python, basicstyle=\ttfamily\scriptsize]
  exist(filter_obj(
      all_boxes, lambda x: x.is_tower() and 
      exist(filter_obj(
          all_boxes, lambda y: 
              y.is_tower() and 
              count(x.all_items_in_box()) == 
                  count(y.all_items_in_box()) and 
              get_set_colors(filter_obj(
                  y.all_items_in_box(), 
                  is_bottom)) != 
              get_set_colors(filter_obj(
                  x.all_items_in_box(), 
                  is_bottom))))))
\end{lstlisting}
\end{minipage}
\vfill

\begin{minipage}{0.4\textwidth}
\begin{tikzpicture}
\node[below=1.3em of TSR0] (TF0)
{\includegraphics[width=1\textwidth]{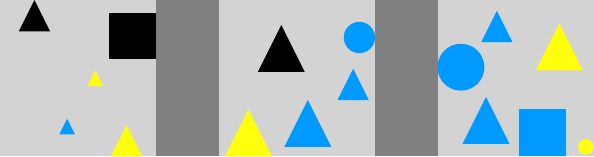}};
\node[align = center, below=-0.3em of TF0, font=\footnotesize\linespread{0.5}] (TFT0) {There is a box with all 3 different colors and \\a black triangle touching the wall with its top.};
\node[rectangle, draw, minimum width=20em, minimum height=12em, fit=(TF0) (TFT0)] (TFR0) {};
\end{tikzpicture}
\end{minipage}
\hfill
\begin{minipage}{0.57\textwidth}
\begin{lstlisting}[language=Python, basicstyle=\ttfamily\scriptsize]
  exist(filter_obj(
      all_boxes, lambda x: 
          count(get_set_colors(
              x.all_items_in_box())) == 3 and 
          exist(filter_obj(
              x.all_items_in_box(), lambda y: 
                  is_black(y) and 
                  is_triangle(y) and 
                  is_touching_wall(y, Side.TOP)))))
\end{lstlisting}
\end{minipage}

\caption{Example sentences with the example images displayed alongside them during annotation (left), and their annotated Python programs (right). Both sentences and programs are \texttt{True} for the corresponding image.} %
\label{fig:logical_forms_side}

\end{figure}

%% file: 120-app_experimental_design.tex
\section{Experimental Setup Details}\label{sec:app:exp}

\subsection{Learning Details}\label{sec:app:exp:learning}

\paragraph{Model Parameters and Computational Resources}\label{sec:app:exp:resources}

\cnnbase and \cnnbaseten use a BERT-base model with 110M parameters. For \vilt,  we use a ViLT-B/32 model with 87.4M parameters (Table 6 in \citet{pmlr-v139-kim21k}).
We use 6 NVIDIA RTX A6000, 3 Titan RTX, and 8 GeForce GTX 2080 Ti for our computations. The total computational budget is 950 GPU hours.

\paragraph{Tokenization}\label{sec:app:exp:tok}

\cnnbase and \cnnbaseten use an uncased BERT WordPiece tokenizer with the default parameters. 
\vilt uses the default  ViLT feature extractor and BERT tokenizer, based on the Hugging Face implementation \citep{wolf-etal-2020-transformers}.

\paragraph{Hyperparameters}
For \cnnbase and \cnnbaseten, we optimize using Adam~\citep{DBLP:journals/corr/KingmaB14}  with a learning rate of 3e-4, except on \toft and on \scft, where we use 3e-5. For \vilt, we use AdamW~\citep{DBLP:conf/iclr/LoshchilovH19} with a cosine scheduler and a base learning rate of 3e-5 for all experiments. The learning rate is warmed up for 1\% of the maximal total training steps of 4M. 
We use patience for early stopping. 
We set entropy to 0{.}1 for all our \envtower experiments and to 0{.}3 for all our \envscatter experiments. 
We use a mini-batch of 64 actions for gradient updates. 
At each \ppo iteration we sample 2{,}048 actions (i.e., for the internal update loop). 

\paragraph{\ppowsf Details}
\ppowsf is a simple variant of \ppo that applies masking to all the actions except for $\stopaction$ when the agent reaches a state in which it will receive a positive reward if it would select $\stopaction$. 
\ppowsf allows the learner to observe $\stopaction$ with positive reward with higher probability than with conventional \ppo. 
A side effect of this masking is that the learner often samples action with very low probability, which can lead to exploding gradients. 
We clip the \ppo ratio to address this.
Formally, the original \ppo objective is:

\begin{small}
\begin{align}
   L(\theta) &= \\\nonumber &\mathbb{E}_t \left[ \min(r_t(\theta)\hat{A}_t, \text{clip}(r_t(\theta), 1-\epsilon, 1+\epsilon) \hat{A}_t\right] \;\;,
\end{align}
\end{small}

\noindent
where $r_t(\theta)=\frac{\pi_\theta(\action_t|\state_t)}{\pi_{old}(\action_t|\state_t)}$, $\hat{A}$ is the advantage function, and $\epsilon$ is a hyperparameter~\citep{schulman2017proximal}. In \ppowsf, we clip the ratio term $r_t(\theta)$ to avoid very large value due to ``force'' sampling of actions with very low probability:

\begin{small}
\begin{equation}
    \hat{r}_{t}(\theta) = \min\biggl(r_t(\theta), M\biggr)\;\;,
\end{equation}
\end{small}

\noindent
where $M$ is a threshold bounding the ratio. We use $\hat{r}_t(\theta)$ in place of $r_t(\theta)$ for our experiments.

\subsection{Inference Details}\label{sec:app:exp:inference}

There are three action types $\stopaction$, $\act{ADD}$, and $\act{REMOVE}$. 
Each type take a different number of arguments: $\stopaction$ takes no arguments, $\act{ADD}$ takes two arguments in \envtower and five in \envscatter, and $\act{REMOVE}$ takes one argument in \envtower and two in \envscatter. 
During inference, actions $\action$ are sampled from the agent policy $\action \sim \pi(\cdot|\state, c)$, where $\state$ is a state and $\context$ is a context. 
We decompose the probability of an action to be a product of its type and arguments. This risks assigning generally lower probability to actions with more arguments, because of the multiplicative decomposition. 
We avoid this by sampling the required arguments as needed. 
We first sample an action type. Depending on the action type, we sample the required arguments. 
In practice, this means that when an argument slot is not used, the probability of that action marginalizes over all possible assignments to that argument slot.

%% file: 130-exp.tex
\section{Additional Results and Analysis}\label{sec:app:results}

\subsection{Development Rollout Statistics}\label{sec:app:results:dev:pposf}

\begin{figure}[t!]
  \includegraphics[width=1\linewidth]{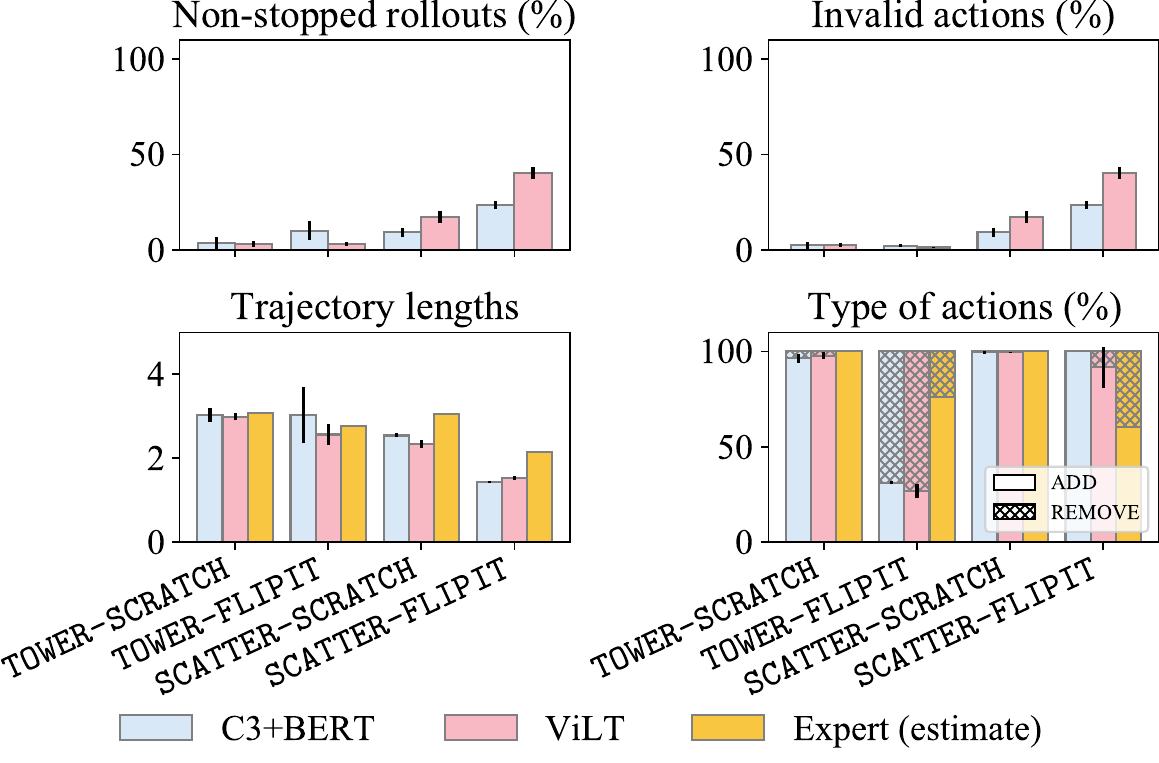}  
  \caption{Mean development statistics for \ppowsf, averaged over three runs. Clockwise from top left: rollouts without a $\stopaction$ action; rollouts with an invalid action; mean actions per rollout; relative rate of $\act{ADD}$/$\act{REMOVE}$ actions.}
  \label{fig:analysis_quant_pposf}
\end{figure}

\autoref{fig:analysis_quant_pposf} shows development rollout statistics for \ppowsf.  The statistics follow similar trends for the ones we show for \ppo in \autoref{fig:analysis_quant}.
Compared to \ppo, we observe more non-stopped rollouts for \toft when training with \ppowsf, and less for \envscatter. 
These non-stopped \toft rollouts often correspond to the model getting stuck in add-remove loops.

\subsection{Error Analysis}\label{sec:app:results:errors}

We analyze model errors by sampling 50 erroneous development examples,\footnote{If there are less than 50 errors in the development set, we analyze the entire set. This occurs only in \scst with \cnnbase.} for the two \envscatter CMDPs trained with \ppo, over one run: 
\begin{description}[leftmargin=.1in]
  \item[\scst with \cnnbase] 58\% of the errors are due to invalid actions, and 42\% due to direct or early termination. Among the invalid actions, all are due to trying to perform an action on a separator. Among the termination errors, 18\% are due to direct termination, and 82\% are due to early termination.
  \item[\scst with \vilt] 82\% of the errors are due to invalid actions, and 18\% due to direct or early termination. Among the invalid actions, 78\% are due to trying to perform an action on a separator, 14\% due to trying to remove an object from a position that does not include an object, 5\% due to trying to put an item that cannot fit in the box, and 3\% due to trying to add an object on top of an existing one.
  Among the termination errors, 50\% are due to direct termination and 50\% due to erroneous termination.
  \item[\scft with \cnnbase] 58\% of the errors are due to invalid actions, and 42\% are due to direct or early termination. 
Among the invalid actions, 63\% are due to trying to remove an object from a position that does not include an object, 24\% are due to trying to perform an action on a separator, 10\% due to trying to put an item that cannot fit in the box, and 3\% due to trying to add an object on top of an existing one.
Among the termination errors, 90\% are due to direct termination and 10\% due to erroneous termination.
  \item[\scft with \vilt] 64\% of the mistakes are due to invalid actions, and 36\% due to early termination. Among the invalid actions, 75\% are due to trying to perform an action on a separator, 19\% due to trying to remove an object from a position that does not include an object, and 6\% due to trying to add an object on top of an existing one.
\end{description}

\subsection{Analysis of Action Selection Bias}\label{sec:app:results:actbias}

\begin{figure}[t]
  \centering
  \scst\\[5pt]
  \includegraphics[width=1\linewidth]{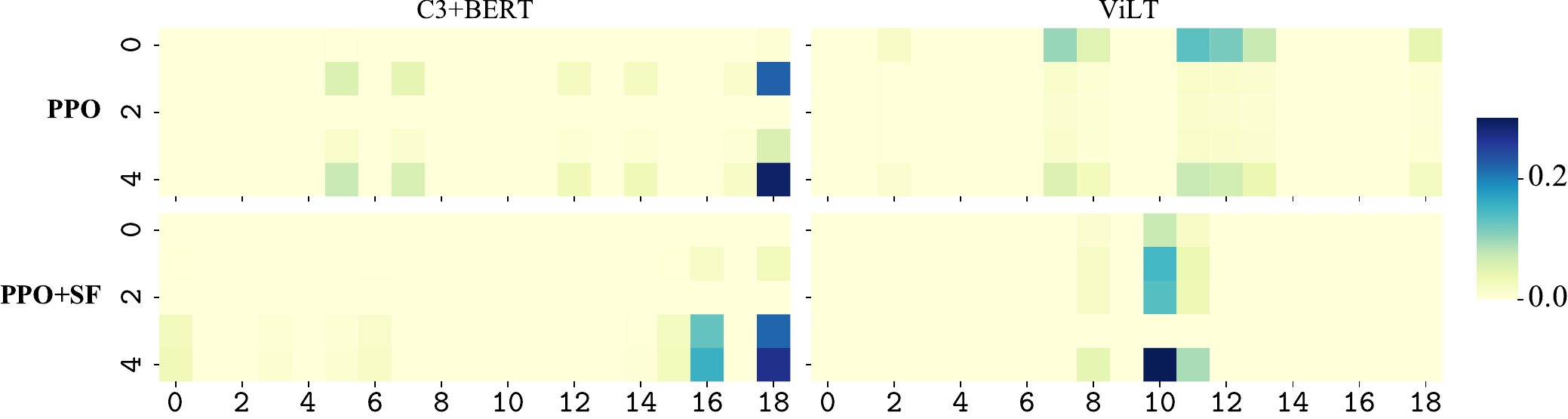} \\[10pt]
  \scft\\[5pt]
\includegraphics[width=1\linewidth]{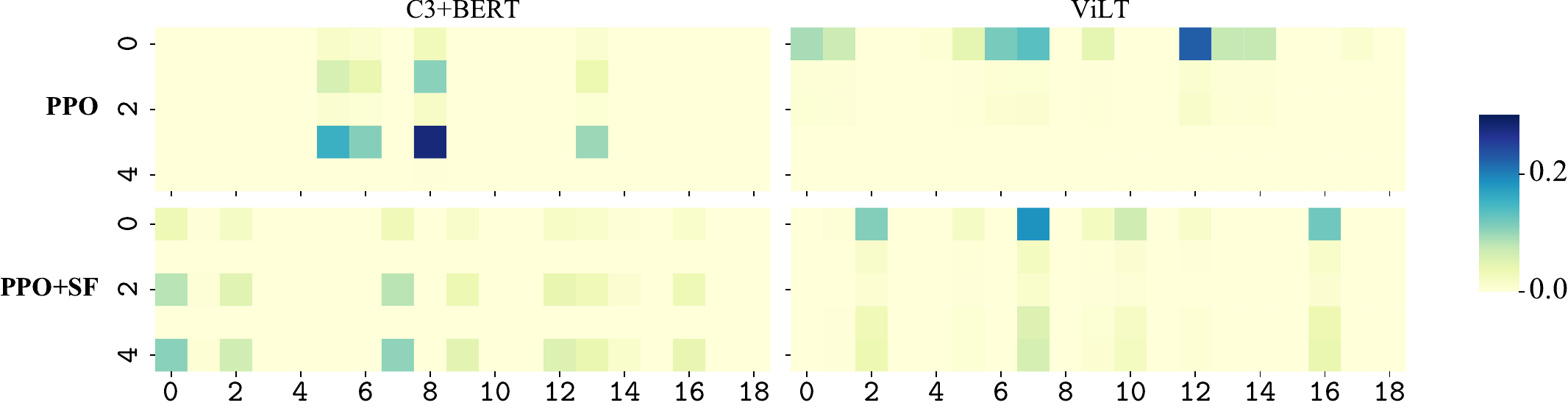}
\caption{Frequency of selecting the grid cell $(x, y)$ in \envscatter CMDPs for rollouts sampled on the development set, over one run. $x$-axis and $y$-axis show the $x$ and $y$ position of the cells in the \envscatter 19$\times$5 grid approximation.}\label{fig:bias:scatter:pos}
\end{figure}

We observe that the trained models often exhibit bias towards specific action arguments, which are sampled much more often than others during inference. 
\autoref{fig:bias:scatter:pos} illustrates this by visualizing coordinate selection frequencies on the development set for \envscatter CMDPs, for one of the runs.
While the presence of bias is relatively persistent, the exact argument the models are biased towards vary.
This indicates generalization limitations of our learned policies, which potentially converge to specific argument prematurely, and do not fully utilize the entire action space. 
We observe that this bias leads to selecting invalid actions, for example when attempting to place a large object on the edge so it crosses image boundaries.

\subsection{Performance Analysis by Semantic and Syntactic Phenomena}\label{sec:app:results:semantics}

\citet{suhr2017corpus} manually annotated 200 development examples for semantic phenomena. 
\autoref{tab:ling_ppo_long} and \autoref{tab:ling_ppo_long_mask} show the performance on this data of policies trained with \ppo and \ppowsf.
We provide an example sentence for each category. 
The two models mostly follow similar trends with respect to the categories on which they perform above and below overall performance.
The two models mostly follow similar trends with respect to the categories on which they perform above and below overall performance. 
When trained with \ppo, both models outperform overall performance on hard cardinality (e.g., \nlstring{\dots exactly four objects \dots}) for \envtower CMDPs, and on presupposition for \taskscratch CMDPs.
On spatial relations, both models perform above overall performance only for \tost, and below for all the other three CMDPs. 
We observe that \ppowsf is especially helpful for this category, bringing the performance of \vilt above average performance on all CMDPs.

%% file: 132-other_models.tex
\input{tables/133-table_results_flava}

\section{Experiments with \flava}\label{sec:app:other_models}

We conduct preliminary experiments with the base \flava model (350M parameters) \citep{singh2022flava}.\footnote{We also experimented with \clip (ViT-B/32) \citep{radford2021learning}, but the performed poorly on the simplest \tost CMDP, so was discarded relatively early.} 
\autoref{tab:results:flava} shows the results.
On \toft, the results with \ppo are outperforming \vilt in \autoref{tab:results}. %
On \tost, with \ppowsf, \flava's results are on par with \vilt, and with \ppo, below \vilt.
On \envscatter environments, \flava's performance is significantly lower than \cnnbase, \cnnbaseten and \vilt in \autoref{tab:results}. 
We tested different hyperparameters, using learning rates from 1e-3 to 3e-6, but di not find a combination that significantly improves the learning behaviour.
Due to the computational resources required in training \flava, and the results on \envtower environments that are comparable but not always outperforming \vilt, we choose to not perform further hyperparameter ssearch on \envscatter.

%% file: tables/133-table_results_flava.tex
\begin{table*}[t]
\centering
\footnotesize
\setlength{\tabcolsep}{2pt}
    \begin{tabular}{lccccccccccc}
    \toprule
    & \multicolumn{2}{c}{\textbf{\tost}} && \multicolumn{2}{c}{\textbf{\toft}} && \multicolumn{2}{c}{\textbf{\scst}} && \multicolumn{2}{c}{\textbf{\scft}}
    \\
    \cmidrule{2-3} \cmidrule{5-6} \cmidrule{8-9} \cmidrule{11-12}
    & Dev & Test && Dev  & Test && Dev & Test && Dev & Test 
    \\ 
    \midrule
     \multirow{1}{*}{\ppo} & 75{.}46  & 65{.}43 && \textbf{62{.}72}  & \textbf{58{.}79} && 12{.}64 & 15{.}48 && \phantom{0}1{.}28  & \phantom{0}0{.}68\\

    \midrule
    
    \multirow{1}{*}{\ppowsf} & \textbf{84{.}05}  & 76{.}24 && 58{.}88  & 58{.}21 && 17{.}24 & 27{.}10 && \phantom{0}7{.}35  & \phantom{0}6{.}94 \\

    \bottomrule
    \end{tabular}
    \caption{Mean task-completion accuracies for \envtower CMDPs using \flava, with seed 1. We optimize using AdamW, and use a learning rate of 3e-5. Bold results are outperforming or on par with \vilt in \autoref{tab:results}.}\label{tab:results:flava}
    \vspace{-10pt}
\end{table*}

%% file: 135-app_third_party.tex
\section{Third-party Code}\label{sec:app:thirdparty}

Whenever the intended use is provided, the use of existing artifacts comply with their intended use. \citet{suhr2017corpus} is under CC-BY-4.0, and \citet{kostrikov2018pytorch} is under MIT. The use of code from \citet{goldman-etal-2018-weakly} was done with explicit approval from the authors, because no license was provided with the code.

%% file: tables/71-ling-analysis_ppo_full.tex
\begin{landscape}
\begin{table}
\centering
\footnotesize
\setlength{\tabcolsep}{2.3pt}
    \begin{tabular}{lcccccccccccl}
    \toprule
    & \multicolumn{2}{c}{\textbf{\tost}} && \multicolumn{2}{c}{\textbf{\toft}} && \multicolumn{2}{c}{\textbf{\scst}} && \multicolumn{2}{c}{\textbf{\scft}} &
    \\
    \cmidrule{2-3} \cmidrule{5-6} \cmidrule{8-9} \cmidrule{11-12}
    & Total & Correct \% && Total  & Correct \% && Total & Correct \% && Total & Correct \% & Example
    \\
    \midrule
    \textbf{Semantics} &&&&&&&&&&& \\
    \midrule
    Cardinality (hard) &98 & \textbf{76{.}5}~{\color{lightgray}\vrule}~\textbf{83{.}7} && 480 & \textbf{28{.}9}~{\color{lightgray}\vrule}~\textbf{56{.}2} && 35 & 49{.}5~{\color{lightgray}\vrule}~40{.}0 && 119 & \phantom{0}0{.}0~{\color{lightgray}\vrule}~\textbf{13{.}7} & \small{\textit{There are \textbf{exactly four objects} not touching any edge}}
    \\
    Cardinality (soft) &21 & 68{.}2~{\color{lightgray}\vrule}~81{.}0 && 82 & \textbf{30{.}1}~{\color{lightgray}\vrule}~\textbf{56{.}5} && 11 & \textbf{60{.}6}~{\color{lightgray}\vrule}~24{.}3 && 42 & \phantom{0}0{.}0~{\color{lightgray}\vrule}~10{.}3 & \small{\textit{There is a box with \textbf{at least one} square and \textbf{at least three} triangles.}}
    \\
    Existential       &122 & \textbf{75{.}1}~{\color{lightgray}\vrule}~\textbf{81{.}7} && 577 & \textbf{29{.}0}~{\color{lightgray}\vrule}~55{.}1 && 55 & \textbf{55{.}7}~{\color{lightgray}\vrule}~35{.}1 && 192 & \phantom{0}0{.}0~{\color{lightgray}\vrule}~12{.}3 & \small{\textit{\textbf{There is a tower} with yellow base.}}
    \\
    Universal &7 & \textbf{85{.}7}~{\color{lightgray}\vrule}~\textbf{95{.}2} && 28 & \textbf{29{.}7}~{\color{lightgray}\vrule}~46{.}4 && 9 & \textbf{81{.}5}~{\color{lightgray}\vrule}~\textbf{59{.}3} && 36 & \phantom{0}0{.}0~{\color{lightgray}\vrule}~\phantom{0}9{.}3 &      \small{\textit{There is a black item in \textbf{every box}.}}
    \\
    Coordination      &19 & \textbf{85{.}9}~{\color{lightgray}\vrule}~\textbf{84{.}2} && 86 & 27{.}1~{\color{lightgray}\vrule}~52{.}7 && 15 & \textbf{64{.}5}~{\color{lightgray}\vrule}~40{.}0 && 55 & \phantom{0}0{.}0~{\color{lightgray}\vrule}~\phantom{0}9{.}7 & \small{\textit{There are 2 blue circles \textbf{and} 1 blue triangle}}
    \\
    Coreference &3 & \textbf{100{.}0}~{\color{lightgray}\vrule}~77{.}8\phantom{0} && 10 & 13{.}3~{\color{lightgray}\vrule}~10{.}0 && 3 & \textbf{44{.}4}~{\color{lightgray}\vrule}~33{.}3 && 9 & \phantom{0}0{.}0~{\color{lightgray}\vrule}~\phantom{0}7{.}4 & \small{\textit{There is a blue triangle touching the wall with \textbf{its} side.}}
    \\
    Spatial Relations &94 & \textbf{74{.}8}~{\color{lightgray}\vrule}~\textbf{81{.}6} && 438 & 26{.}6~{\color{lightgray}\vrule}~53{.}1 && 39 & 53{.}8~{\color{lightgray}\vrule}~32{.}5 && 128 & \phantom{0}0{.}0~{\color{lightgray}\vrule}~10{.}1 &       \small{\textit{there is one tower with a yellow block \textbf{above} a yellow block}}
    \\
    Comparative &5 & 66{.}7~{\color{lightgray}\vrule}~73{.}3 && 20 & 11{.}7~{\color{lightgray}\vrule}~21{.}7 && 1 & \textbf{100{.}0}~{\color{lightgray}\vrule}~\textbf{100{.}0} && 4 & \phantom{0}0{.}0~{\color{lightgray}\vrule}~\textbf{16{.}7} & \small{\textit{There is a box with multiple items and only one item \textbf{has a different color.}}}
    \\
    Presupposition    &17 & \textbf{74{.}5}~{\color{lightgray}\vrule}~\textbf{90{.}2} && 74 & 27{.}0~{\color{lightgray}\vrule}~\textbf{54{.}1} && 22 & \textbf{66{.}7}~{\color{lightgray}\vrule}~\textbf{51{.}5} && 78 & \phantom{0}0{.}0~{\color{lightgray}\vrule}~12{.}4 &    \small{\textit{There is a box with seven items and \textbf{the three black items}
are the same in shape.}}
    \\
    Negation &4 & \textbf{75{.}0}~{\color{lightgray}\vrule}~66{.}7 && 15 & 13{.}3~{\color{lightgray}\vrule}~37{.}8 && 14 & 54{.}8~{\color{lightgray}\vrule}~33{.}3 && 52 & \phantom{0}0{.}0~{\color{lightgray}\vrule}~\phantom{0}7{.}0 & \small{\textit{there is exactly one black triangle \textbf{not touching} the edge}}
    \\
    \midrule
    \textbf{Syntax} &&&&&&&&&&& \\
    \midrule
    Coordination &4 & \textbf{83{.}3}~{\color{lightgray}\vrule}~75{.}0 && 14 & 11{.}9~{\color{lightgray}\vrule}~\textbf{59{.}5} && 5 & 53{.}3~{\color{lightgray}\vrule}~26{.}7 && 20 & \phantom{0}0{.}0~{\color{lightgray}\vrule}~\phantom{0}8{.}3 &         \small{\textit{There is a box with at least one square \textbf{and} at least three triangles.}}
    \\
    PP Attachment &44 & \textbf{76{.}5}~{\color{lightgray}\vrule}~\textbf{81{.}8} && 215 & 26{.}2~{\color{lightgray}\vrule}~\textbf{54{.}3} && 3 & 33{.}3~{\color{lightgray}\vrule}~33{.}3 && 8 & \phantom{0}0{.}0~{\color{lightgray}\vrule}~\phantom{0}8{.}3 &
            \small{\textit{There is a black block on a black block as the base of a tower \textbf{with} three blocks.}}
    \\
    \midrule
    \multicolumn{2}{l}{\textbf{Overall}} & 72{.}80~{\color{lightgray}\vrule}~81{.}19 && & 28{.}11~{\color{lightgray}\vrule}~53{.}25 &&  & 59{.}00~{\color{lightgray}\vrule}~40{.}23 &&  & \phantom{0}0{.}00~{\color{lightgray}\vrule}~13{.}31 \\
    \bottomrule
    \end{tabular}
\caption{Performance on a set of development examples annotated for semantic and syntactic categories by \citet{suhr2017corpus} for both models (\cnnbase~|~\vilt) when trained with \ppo. 
Dev performance refers to mean performance on the respective full development set. Results outperforming dev performance are in bold.}\label{tab:ling_ppo_long}
\end{table}
\end{landscape}

%% file: tables/71-ling-analysis_ppo_sf_full.tex
\begin{landscape}
\begin{table}
\centering
\footnotesize
\setlength{\tabcolsep}{2.3pt}
    \begin{tabular}{lcccccccccccl}
    \toprule
    & \multicolumn{2}{c}{\textbf{\tost}} && \multicolumn{2}{c}{\textbf{\toft}} && \multicolumn{2}{c}{\textbf{\scst}} && \multicolumn{2}{c}{\textbf{\scft}} &
    \\
    \cmidrule{2-3} \cmidrule{5-6} \cmidrule{8-9} \cmidrule{11-12}
    & Total & Correct \% && Total  & Correct \% && Total & Correct \% && Total & Correct \% & Example
    \\
    \midrule
    \textbf{Semantics} &&&&&&&&&&& \\
    \midrule
    Cardinality (hard) &98 & \textbf{84{.}3}~{\color{lightgray}\vrule}~\textbf{85{.}1} && 480 & 30{.}3~{\color{lightgray}\vrule}~\textbf{68{.}9} && 35 & 60{.}9~{\color{lightgray}\vrule}~59{.}1 && 119 & \textbf{17{.}1}~{\color{lightgray}\vrule}~27{.}7 &  \small{\textit{There are \textbf{exactly four objects} not touching any edge}}
    
    \\
    Cardinality (soft)&21 & 76{.}2~{\color{lightgray}\vrule}~\textbf{87{.}3} && 82 & \textbf{33{.}7}~{\color{lightgray}\vrule}~\textbf{66{.}7} && 11 & \textbf{72{.}7}~{\color{lightgray}\vrule}~\textbf{78{.}8} && 42 & 10{.}3~{\color{lightgray}\vrule}~18{.}3 &
 \small{\textit{There is a box with \textbf{at least one} square and \textbf{at least three} triangles.}}
    \\
    Existential     &122 & \textbf{84{.}1}~{\color{lightgray}\vrule}~83{.}6 && 577 & 31{.}8~{\color{lightgray}\vrule}~\textbf{69{.}5} && 55 & 70{.}3~{\color{lightgray}\vrule}~67{.}3 && 192 & 15{.}8~{\color{lightgray}\vrule}~27{.}8 &
 \small{\textit{\textbf{There is a tower} with yellow base.}}
    \\
    Universal &7 & 80{.}9~{\color{lightgray}\vrule}~\textbf{95{.}2} && 28 & 26{.}2~{\color{lightgray}\vrule}~52{.}4 && 9 & \textbf{88{.}9}~{\color{lightgray}\vrule}~\textbf{88{.}9} && 36 & \phantom{0}7{.}4~{\color{lightgray}\vrule}~16{.}6 &\small{\textit{There is a black item in \textbf{every box}.}}
    \\
    Coordination    &19 & \textbf{91{.}2}~{\color{lightgray}\vrule}~\textbf{84{.}2} && 86 & \textbf{39{.}9}~{\color{lightgray}\vrule}~65{.}1 && 15 & \textbf{73{.}3}~{\color{lightgray}\vrule}~64{.}5 && 55 & \phantom{0}6{.}1~{\color{lightgray}\vrule}~17{.}0 & \small{\textit{There are 2 blue circles \textbf{and} 1 blue triangle}}
    \\
    Coreference &3 & \textbf{\phantom{0}88{.}9}~{\color{lightgray}\vrule}~\textbf{100{.}0} && 10 & 23{.}3~{\color{lightgray}\vrule}~30{.}0 && 3 & 44{.}4~{\color{lightgray}\vrule}~44{.}4 && 9 & \phantom{0}3{.}7~{\color{lightgray}\vrule}~\textbf{33{.}3} & \small{\textit{There is a blue triangle touching the wall with \textbf{its} side.}}
    \\
    Spatial Relations &94 & \textbf{81{.}9}~{\color{lightgray}\vrule}~\textbf{84{.}4} && 438 & 26{.}9~{\color{lightgray}\vrule}~\textbf{69{.}0} && 39 & 68{.}4~{\color{lightgray}\vrule}~\textbf{68{.}4} && 128 & 16{.}7~{\color{lightgray}\vrule}~\textbf{28{.}1} &       \small{\textit{there is one tower with a yellow block \textbf{above} a yellow block}}
    \\
    Comparative &5 & 73{.}3~{\color{lightgray}\vrule}~80{.}0 && 20 & 21{.}7~{\color{lightgray}\vrule}~31{.}7 && 1 & \textbf{100{.}0}~{\color{lightgray}\vrule}~\textbf{100{.}0} && 4 & 16{.}7~{\color{lightgray}\vrule}~\textbf{41{.}7} & \small{\textit{There is a box with multiple items and only one item \textbf{has a different color.}}}
    \\
    Presupposition   &17 & \textbf{82{.}4}~{\color{lightgray}\vrule}~\textbf{94{.}1} && 74 & 29{.}3~{\color{lightgray}\vrule}~65{.}3 && 22 & \textbf{72{.}7}~{\color{lightgray}\vrule}~\textbf{69{.}7} && 78 & \textbf{14{.}1}~{\color{lightgray}\vrule}~\textbf{29{.}9} &    \small{\textit{There is a box with seven items and \textbf{the three black items}
are the same in shape.}}
    \\
    Negation &4 & 75{.}0~{\color{lightgray}\vrule}~58{.}3 && 15 & 15{.}6~{\color{lightgray}\vrule}~64{.}4 && 14 & 66{.}7~{\color{lightgray}\vrule}~\textbf{71{.}4} && 52 & \textbf{17{.}3}~{\color{lightgray}\vrule}~23{.}7 & \small{\textit{there is exactly one black triangle \textbf{not touching} the edge}}
    \\
    \midrule
    \textbf{Syntax} &&&&&&&&&&& \\
    \midrule
    Coordination &4 & \textbf{91{.}7}~{\color{lightgray}\vrule}~75{.}0 && 14 & 19{.}0~{\color{lightgray}\vrule}~\textbf{69{.}0} && 5 & 60{.}0~{\color{lightgray}\vrule}~46{.}7 && 20 & \phantom{0}5{.}0~{\color{lightgray}\vrule}~20{.}0 &        \small{\textit{There is a box with at least one square \textbf{and} at least three triangles.}}
    \\
    PP Attachment &44 & \textbf{84{.}8}~{\color{lightgray}\vrule}~\textbf{85{.}6} && 215 & 27{.}0~{\color{lightgray}\vrule}~\textbf{70{.}2} && 3 & \textbf{77{.}8}~{\color{lightgray}\vrule}~66{.}7 && 8 & \phantom{0}8{.}3~{\color{lightgray}\vrule}~20{.}8 &        \small{\textit{There is a black block on a black block as the base of a tower \textbf{with} three blocks.}}
    \\
    \midrule
    \multicolumn{2}{l}{\textbf{Overall}} & 81{.}80~{\color{lightgray}\vrule}~84{.}05 && & 32{.}59~{\color{lightgray}\vrule}~65{.}68 &&  & 72{.}03~{\color{lightgray}\vrule}~67{.}43 &&  & 17{.}04~{\color{lightgray}\vrule}~28{.}01 \\
    \bottomrule
    \end{tabular}
\caption{Performance on a set of development examples annotated for semantic and syntactic categories by \citet{suhr2017corpus} for both models (\cnnbase~|~\vilt) when trained with \ppowsf. 
Dev performance refers to mean performance on the respective full development set. Results outperforming dev performance are in bold.}\label{tab:ling_ppo_long_mask}
\end{table}
\end{landscape}

%% file: acl2023.bbl
\begin{thebibliography}{47}
\expandafter\ifx\csname natexlab\endcsname\relax\def\natexlab#1{#1}\fi

\bibitem[{Anderson et~al.(2018)Anderson, Wu, Teney, Bruce, Johnson,
  S{\"u}nderhauf, Reid, Gould, and van~den Hengel}]{Anderson:18r2r}
Peter Anderson, Qi~Wu, Damien Teney, Jake Bruce, Mark Johnson, Niko
  S{\"u}nderhauf, Ian Reid, Stephen Gould, and Anton van~den Hengel. 2018.
\newblock Vision-and-language navigation: Interpreting visually-grounded
  navigation instructions in real environments.
\newblock In \emph{The IEEE Conference on Computer Vision and Pattern
  Recognition}, pages 3674--3683.

\bibitem[{Bellemare et~al.(2013)Bellemare, Naddaf, Veness, and
  Bowling}]{bellemare2013arcade}
Marc~G Bellemare, Yavar Naddaf, Joel Veness, and Michael Bowling. 2013.
\newblock The arcade learning environment: An evaluation platform for general
  agents.
\newblock \emph{Journal of Artificial Intelligence Research}, 47:253--279.

\bibitem[{Brockman et~al.(2016)Brockman, Cheung, Pettersson, Schneider,
  Schulman, Tang, and Zaremba}]{Brockman2016:openai-gym}
Greg Brockman, Vicki Cheung, Ludwig Pettersson, Jonas Schneider, John Schulman,
  Jie Tang, and Wojciech Zaremba. 2016.
\newblock Openai gym.
\newblock \emph{arXiv preprint arXiv:1606.01540}.

\bibitem[{Cao et~al.(2020)Cao, Wang, Zhang, and
  Manivasagam}]{DBLP:journals/corr/abs-2004-07200}
Tianshi Cao, Jingkang Wang, Yining Zhang, and Sivabalan Manivasagam. 2020.
\newblock Babyai++: Towards grounded-language learning beyond memorization.
\newblock \emph{Beyond tabula rasa in RL (BeTR-RL) Workshop held in conjunction
  with the 8th International Conference on Learning Representations, {ICLR}}.

\bibitem[{Chen et~al.(2019)Chen, Suhr, Misra, Snavely, and
  Artzi}]{chen2019touchdown}
Howard Chen, Alane Suhr, Dipendra Misra, Noah Snavely, and Yoav Artzi. 2019.
\newblock Touchdown: Natural language navigation and spatial reasoning in
  visual street environments.
\newblock In \emph{Proceedings of the IEEE/CVF Conference on Computer Vision
  and Pattern Recognition}, pages 12538--12547.

\bibitem[{Chevalier{-}Boisvert et~al.(2019)Chevalier{-}Boisvert, Bahdanau,
  Lahlou, Willems, Saharia, Nguyen, and
  Bengio}]{DBLP:conf/iclr/Chevalier-Boisvert19}
Maxime Chevalier{-}Boisvert, Dzmitry Bahdanau, Salem Lahlou, Lucas Willems,
  Chitwan Saharia, Thien~Huu Nguyen, and Yoshua Bengio. 2019.
\newblock Babyai: {A} platform to study the sample efficiency of grounded
  language learning.
\newblock In \emph{7th International Conference on Learning Representations,
  {ICLR}}.

\bibitem[{Co-Reyes et~al.(2019)Co-Reyes, Gupta, Sanjeev, Altieri, DeNero,
  Abbeel, and Levine}]{CoReyes2019:language-meta-rl}
John~D. Co-Reyes, Abhishek Gupta, Suvansh Sanjeev, Nick Altieri, John DeNero,
  P.~Abbeel, and Sergey Levine. 2019.
\newblock Guiding policies with language via meta-learning.
\newblock In \emph{7th International Conference on Learning Representations,
  {ICLR}}.

\bibitem[{C{\^{o}}t{\'{e}} et~al.(2018)C{\^{o}}t{\'{e}}, K{\'{a}}d{\'{a}}r,
  Yuan, Kybartas, Barnes, Fine, Moore, Hausknecht, Asri, Adada, Tay, and
  Trischler}]{DBLP:conf/ijcai/CoteKYKBFMHAATT18}
Marc{-}Alexandre C{\^{o}}t{\'{e}}, {\'{A}}kos K{\'{a}}d{\'{a}}r, Xingdi Yuan,
  Ben Kybartas, Tavian Barnes, Emery Fine, James Moore, Matthew~J. Hausknecht,
  Layla~El Asri, Mahmoud Adada, Wendy Tay, and Adam Trischler. 2018.
\newblock Textworld: {A} learning environment for text-based games.
\newblock In \emph{Computer Games Workshop (CGW) held in conjunction with the
  27th International Conference on Artificial Intelligence, {IJCAI}}.

\bibitem[{Dasigi et~al.(2019)Dasigi, Gardner, Murty, Zettlemoyer, and
  Hovy}]{dasigi-etal-2019-iterative}
Pradeep Dasigi, Matt Gardner, Shikhar Murty, Luke Zettlemoyer, and Eduard Hovy.
  2019.
\newblock Iterative search for weakly supervised semantic parsing.
\newblock In \emph{Proceedings of the 2019 Conference of the North {A}merican
  Chapter of the Association for Computational Linguistics: Human Language
  Technologies, Volume 1 (Long and Short Papers)}, pages 2669--2680.

\bibitem[{Devlin et~al.(2019)Devlin, Chang, Lee, and
  Toutanova}]{devlin-etal-2019-bert}
Jacob Devlin, Ming-Wei Chang, Kenton Lee, and Kristina Toutanova. 2019.
\newblock {BERT}: Pre-training of deep bidirectional transformers for language
  understanding.
\newblock In \emph{Proceedings of the 2019 Conference of the North {A}merican
  Chapter of the Association for Computational Linguistics: Human Language
  Technologies, Volume 1 (Long and Short Papers)}, pages 4171--4186.

\bibitem[{Fukushima and Miyake(1982)}]{fukushima1982neocognitron}
Kunihiko Fukushima and Sei Miyake. 1982.
\newblock Neocognitron: A self-organizing neural network model for a mechanism
  of visual pattern recognition.
\newblock In \emph{Competition and Cooperation in Neural Nets}, pages 267--285.
  Springer.

\bibitem[{Goldman et~al.(2018)Goldman, Latcinnik, Nave, Globerson, and
  Berant}]{goldman-etal-2018-weakly}
Omer Goldman, Veronica Latcinnik, Ehud Nave, Amir Globerson, and Jonathan
  Berant. 2018.
\newblock Weakly supervised semantic parsing with abstract examples.
\newblock In \emph{Proceedings of the 56th Annual Meeting of the Association
  for Computational Linguistics (Volume 1: Long Papers)}, pages 1809--1819.

\bibitem[{Gupta et~al.(2021)Gupta, Singh, and
  Gardner}]{gupta-etal-2021-enforcing}
Nitish Gupta, Sameer Singh, and Matt Gardner. 2021.
\newblock Enforcing consistency in weakly supervised semantic parsing.
\newblock In \emph{Proceedings of the 59th Annual Meeting of the Association
  for Computational Linguistics and the 11th International Joint Conference on
  Natural Language Processing (Volume 2: Short Papers)}, pages 168--174.

\bibitem[{Hallak et~al.(2015)Hallak, Di~Castro, and
  Mannor}]{hallak2015contextual}
Assaf Hallak, Dotan Di~Castro, and Shie Mannor. 2015.
\newblock Contextual markov decision processes.
\newblock \emph{arXiv preprint arXiv:1502.02259}.

\bibitem[{Hanjie et~al.(2021)Hanjie, Zhong, and
  Narasimhan}]{hanjie2021grounding}
Austin~W Hanjie, Victor~Y Zhong, and Karthik Narasimhan. 2021.
\newblock Grounding language to entities and dynamics for generalization in
  reinforcement learning.
\newblock In \emph{International Conference on Machine Learning}, pages
  4051--4062.

\bibitem[{Hausknecht et~al.(2020)Hausknecht, Ammanabrolu, C{\^{o}}t{\'{e}}, and
  Yuan}]{DBLP:conf/aaai/HausknechtACY20}
Matthew~J. Hausknecht, Prithviraj Ammanabrolu, Marc{-}Alexandre
  C{\^{o}}t{\'{e}}, and Xingdi Yuan. 2020.
\newblock Interactive fiction games: {A} colossal adventure.
\newblock In \emph{Proceedings of the AAAI Conference on Artificial
  Intelligence}, volume~34.

\bibitem[{Hermann et~al.(2017)Hermann, Hill, Green, Wang, Faulkner, Soyer,
  Szepesvari, Czarnecki, Jaderberg, Teplyashin, Wainwright, Apps, Hassabis, and
  Blunsom}]{DBLP:journals/corr/HermannHGWFSSCJ17}
Karl~Moritz Hermann, Felix Hill, Simon Green, Fumin Wang, Ryan Faulkner, Hubert
  Soyer, David Szepesvari, Wojciech~Marian Czarnecki, Max Jaderberg, Denis
  Teplyashin, Marcus Wainwright, Chris Apps, Demis Hassabis, and Phil Blunsom.
  2017.
\newblock Grounded language learning in a simulated 3d world.
\newblock \emph{arXiv preprint arXiv:1706.06551}.

\bibitem[{Hudson and Manning(2018)}]{DBLP:conf/iclr/HudsonM18}
Drew~A. Hudson and Christopher~D. Manning. 2018.
\newblock Compositional attention networks for machine reasoning.
\newblock In \emph{6th International Conference on Learning Representations,
  {ICLR}}.

\bibitem[{Jiang et~al.(2020)Jiang, Luketina, Nardelli, Minervini, Torr,
  Whiteson, and Rockt{\"a}schel}]{jiang2020wordcraft}
Minqi Jiang, Jelena Luketina, Nantas Nardelli, Pasquale Minervini, Philip~HS
  Torr, Shimon Whiteson, and Tim Rockt{\"a}schel. 2020.
\newblock Wordcraft: An environment for benchmarking commonsense agents.
\newblock \emph{arXiv preprint arXiv:2007.09185}.

\bibitem[{Johnson et~al.(2017)Johnson, Hariharan, van~der Maaten, Hoffman,
  Fei-Fei, Zitnick, and Girshick}]{Johnson2017:symbolic-clevr}
Justin Johnson, Bharath Hariharan, Laurens van~der Maaten, Judy Hoffman,
  Li~Fei-Fei, C.~Lawrence Zitnick, and Ross~B. Girshick. 2017.
\newblock Inferring and executing programs for visual reasoning.
\newblock \emph{2017 IEEE International Conference on Computer Vision (ICCV)},
  pages 3008--3017.

\bibitem[{Kim et~al.(2021)Kim, Son, and Kim}]{pmlr-v139-kim21k}
Wonjae Kim, Bokyung Son, and Ildoo Kim. 2021.
\newblock Vilt: Vision-and-language transformer without convolution or region
  supervision.
\newblock In \emph{Proceedings of the 38th International Conference on Machine
  Learning}, pages 5583--5594.

\bibitem[{Kingma and Ba(2015)}]{DBLP:journals/corr/KingmaB14}
Diederik~P. Kingma and Jimmy Ba. 2015.
\newblock Adam: {A} method for stochastic optimization.
\newblock In \emph{3rd International Conference on Learning Representations,
  {ICLR}}.

\bibitem[{Kostrikov(2018)}]{kostrikov2018pytorch}
Ilya Kostrikov. 2018.
\newblock Pytorch implementations of reinforcement learning algorithms.

\bibitem[{Ku et~al.(2020)Ku, Anderson, Patel, Ie, and
  Baldridge}]{Ku2020:room-across-room}
Alexander Ku, Peter Anderson, Roma Patel, Eugene Ie, and Jason Baldridge. 2020.
\newblock Room-across-room: Multilingual vision-and-language navigation with
  dense spatiotemporal grounding.
\newblock In \emph{Proceedings of the 2020 Conference on Empirical Methods in
  Natural Language Processing (EMNLP)}, pages 4392--4412.

\bibitem[{Loshchilov and Hutter(2019)}]{DBLP:conf/iclr/LoshchilovH19}
Ilya Loshchilov and Frank Hutter. 2019.
\newblock Decoupled weight decay regularization.
\newblock In \emph{7th International Conference on Learning Representations,
  {ICLR}}.

\bibitem[{Mnih et~al.(2013)Mnih, Kavukcuoglu, Silver, Graves, Antonoglou,
  Wierstra, and Riedmiller}]{mnih2013playing}
Volodymyr Mnih, Koray Kavukcuoglu, David Silver, Alex Graves, Ioannis
  Antonoglou, Daan Wierstra, and Martin Riedmiller. 2013.
\newblock Playing atari with deep reinforcement learning.
\newblock \emph{arXiv preprint arXiv:1312.5602}.

\bibitem[{Narasimhan et~al.(2015)Narasimhan, Kulkarni, and
  Barzilay}]{DBLP:conf/emnlp/NarasimhanKB15}
Karthik Narasimhan, Tejas~D. Kulkarni, and Regina Barzilay. 2015.
\newblock Language understanding for text-based games using deep reinforcement
  learning.
\newblock In \emph{Proceedings of the 2015 Conference on Empirical Methods in
  Natural Language Processing, {EMNLP}}, pages 1--11.

\bibitem[{Norman et~al.(2021)Norman, Kellen, Smallen, DeMeulle, Strande,
  Lazowska, Alterman, Fatland, Stone, Tan, Yelick, Van~Dusen, and
  Mitchell}]{Norman2021:cloudbank}
Michael Norman, Vince Kellen, Shava Smallen, Brian DeMeulle, Shawn Strande,
  Ed~Lazowska, Naomi Alterman, Rob Fatland, Sarah Stone, Amanda Tan, Katherine
  Yelick, Eric Van~Dusen, and James Mitchell. 2021.
\newblock Cloudbank: Managed services to simplify cloud access for computer
  science research and education.
\newblock In \emph{Practice and Experience in Advanced Research Computing},
  PEARC '21. Association for Computing Machinery.

\bibitem[{Pavez et~al.(2018)Pavez, Allende, and
  Allende-Cid}]{pavez-etal-2018-working}
Juan Pavez, H{\'e}ctor Allende, and H{\'e}ctor Allende-Cid. 2018.
\newblock Working memory networks: Augmenting memory networks with a relational
  reasoning module.
\newblock In \emph{Proceedings of the 56th Annual Meeting of the Association
  for Computational Linguistics (Volume 1: Long Papers)}, pages 1000--1009.

\bibitem[{Perez et~al.(2018)Perez, Strub, De~Vries, Dumoulin, and
  Courville}]{perez2018film}
Ethan Perez, Florian Strub, Harm De~Vries, Vincent Dumoulin, and Aaron
  Courville. 2018.
\newblock Film: Visual reasoning with a general conditioning layer.
\newblock In \emph{Proceedings of the AAAI Conference on Artificial
  Intelligence}, volume~32.

\bibitem[{Radford et~al.(2021)Radford, Kim, Hallacy, Ramesh, Goh, Agarwal,
  Sastry, Askell, Mishkin, Clark et~al.}]{radford2021learning}
Alec Radford, Jong~Wook Kim, Chris Hallacy, Aditya Ramesh, Gabriel Goh,
  Sandhini Agarwal, Girish Sastry, Amanda Askell, Pamela Mishkin, Jack Clark,
  et~al. 2021.
\newblock Learning transferable visual models from natural language
  supervision.
\newblock In \emph{International Conference on Machine Learning}, pages
  8748--8763.

\bibitem[{Schulman et~al.(2017)Schulman, Wolski, Dhariwal, Radford, and
  Klimov}]{schulman2017proximal}
John Schulman, Filip Wolski, Prafulla Dhariwal, Alec Radford, and Oleg Klimov.
  2017.
\newblock Proximal policy optimization algorithms.
\newblock \emph{arXiv preprint arXiv:1707.06347}.

\bibitem[{Shridhar et~al.(2020)Shridhar, Thomason, Gordon, Bisk, Han, Mottaghi,
  Zettlemoyer, and Fox}]{Shridhar2020:alfred}
Mohit Shridhar, Jesse Thomason, Daniel Gordon, Yonatan Bisk, Winson Han,
  Roozbeh Mottaghi, Luke Zettlemoyer, and Dieter Fox. 2020.
\newblock {ALFRED}: A benchmark for interpreting grounded instructions for
  everyday tasks.
\newblock In \emph{2020 {IEEE/CVF} Conference on Computer Vision and Pattern
  Recognition, {CVPR}}, pages 10740--10749.

\bibitem[{Shridhar et~al.(2021)Shridhar, Yuan, C{\^{o}}t{\'{e}}, Bisk,
  Trischler, and Hausknecht}]{DBLP:conf/iclr/ShridharYCBTH21}
Mohit Shridhar, Xingdi Yuan, Marc{-}Alexandre C{\^{o}}t{\'{e}}, Yonatan Bisk,
  Adam Trischler, and Matthew~J. Hausknecht. 2021.
\newblock Alfworld: Aligning text and embodied environments for interactive
  learning.
\newblock In \emph{9th International Conference on Learning Representations,
  {ICLR}}.

\bibitem[{Singh et~al.(2022)Singh, Hu, Goswami, Couairon, Galuba, Rohrbach, and
  Kiela}]{singh2022flava}
Amanpreet Singh, Ronghang Hu, Vedanuj Goswami, Guillaume Couairon, Wojciech
  Galuba, Marcus Rohrbach, and Douwe Kiela. 2022.
\newblock Flava: A foundational language and vision alignment model.
\newblock In \emph{Proceedings of the IEEE/CVF Conference on Computer Vision
  and Pattern Recognition}, pages 15638--15650.

\bibitem[{Suhr et~al.(2018)Suhr, Iyer, and Artzi}]{Suhr:18context}
Alane Suhr, Srinivasan Iyer, and Yoav Artzi. 2018.
\newblock Learning to map context-dependent sentences to executable formal
  queries.
\newblock In \emph{Proceedings of the Conference of the North American Chapter
  of the Association for Computational Linguistics: Human Language
  Technologies}, pages 2238--2249.

\bibitem[{Suhr et~al.(2017)Suhr, Lewis, Yeh, and Artzi}]{suhr2017corpus}
Alane Suhr, Mike Lewis, James Yeh, and Yoav Artzi. 2017.
\newblock A corpus of natural language for visual reasoning.
\newblock In \emph{Proceedings of the 55th Annual Meeting of the Association
  for Computational Linguistics (Volume 2: Short Papers)}, pages 217--223.

\bibitem[{Suhr et~al.(2019)Suhr, Zhou, Zhang, Zhang, Bai, and
  Artzi}]{suhr-etal-2019-corpus}
Alane Suhr, Stephanie Zhou, Ally Zhang, Iris Zhang, Huajun Bai, and Yoav Artzi.
  2019.
\newblock A corpus for reasoning about natural language grounded in
  photographs.
\newblock In \emph{Proceedings of the 57th Annual Meeting of the Association
  for Computational Linguistics}, pages 6418--6428.

\bibitem[{Tan and Bansal(2018)}]{tan-bansal-2018-object}
Hao Tan and Mohit Bansal. 2018.
\newblock Object ordering with bidirectional matchings for visual reasoning.
\newblock In \emph{Proceedings of the 2018 Conference of the North {A}merican
  Chapter of the Association for Computational Linguistics: Human Language
  Technologies, Volume 2 (Short Papers)}, pages 444--451.

\bibitem[{Todorov et~al.(2012)Todorov, Erez, and Tassa}]{todorov2012mujoco}
Emanuel Todorov, Tom Erez, and Yuval Tassa. 2012.
\newblock Mujoco: A physics engine for model-based control.
\newblock In \emph{2012 IEEE/RSJ International Conference on Intelligent Robots
  and Systems}, pages 5026--5033.

\bibitem[{Urbanek et~al.(2019)Urbanek, Fan, Karamcheti, Jain, Humeau, Dinan,
  Rockt{\"{a}}schel, Kiela, Szlam, and
  Weston}]{DBLP:conf/emnlp/UrbanekFKJHDRKS19}
Jack Urbanek, Angela Fan, Siddharth Karamcheti, Saachi Jain, Samuel Humeau,
  Emily Dinan, Tim Rockt{\"{a}}schel, Douwe Kiela, Arthur Szlam, and Jason
  Weston. 2019.
\newblock Learning to speak and act in a fantasy text adventure game.
\newblock In \emph{Proceedings of the 2019 Conference on Empirical Methods in
  Natural Language Processing and the 9th International Joint Conference on
  Natural Language Processing, {EMNLP-IJCNLP} 2019}, pages 673--683.

\bibitem[{Wolf et~al.(2020)Wolf, Debut, Sanh, Chaumond, Delangue, Moi, Cistac,
  Rault, Louf, Funtowicz, Davison, Shleifer, von Platen, Ma, Jernite, Plu, Xu,
  Scao, Gugger, Drame, Lhoest, and Rush}]{wolf-etal-2020-transformers}
Thomas Wolf, Lysandre Debut, Victor Sanh, Julien Chaumond, Clement Delangue,
  Anthony Moi, Pierric Cistac, Tim Rault, Rémi Louf, Morgan Funtowicz, Joe
  Davison, Sam Shleifer, Patrick von Platen, Clara Ma, Yacine Jernite, Julien
  Plu, Canwen Xu, Teven~Le Scao, Sylvain Gugger, Mariama Drame, Quentin Lhoest,
  and Alexander~M. Rush. 2020.
\newblock Transformers: State-of-the-art natural language processing.
\newblock In \emph{Proceedings of the 2020 Conference on Empirical Methods in
  Natural Language Processing: System Demonstrations}, pages 38--45.

\bibitem[{Wong et~al.(2021)Wong, Ellis, Tenenbaum, and
  Andreas}]{Wong2021:language-prog-synthesis}
Catherine Wong, Kevin Ellis, Joshua~B. Tenenbaum, and Jacob Andreas. 2021.
\newblock Leveraging language to learn program abstractions and search
  heuristics.
\newblock In \emph{Proceedings of the 38th International Conference on Machine
  Learning, {ICML}}, pages 11193--11204.

\bibitem[{Yao et~al.(2018)Yao, Xu, Wang, and Xu}]{yao-etal-2018-cascaded}
Yiqun Yao, Jiaming Xu, Feng Wang, and Bo~Xu. 2018.
\newblock Cascaded mutual modulation for visual reasoning.
\newblock In \emph{Proceedings of the 2018 Conference on Empirical Methods in
  Natural Language Processing}, pages 975--980.

\bibitem[{Zelle and Mooney(1996)}]{Zelle:96}
J.M. Zelle and Raymond~J. Mooney. 1996.
\newblock Learning to parse database queries using inductive logic programming.
\newblock In \emph{Proceedings of the National Conference on Artificial
  Intelligence}.

\bibitem[{Zettlemoyer and Collins(2005)}]{Zettlemoyer:05}
Luke~S. Zettlemoyer and Michael Collins. 2005.
\newblock Learning to map sentences to logical form: Structured classification
  with probabilistic categorial grammars.
\newblock In \emph{Proceedings of the Conference on Uncertainty in Artificial
  Intelligence}.

\bibitem[{Zheng et~al.(2020)Zheng, Yan, Gou, and
  Wang}]{Zheng2020:webly-sup-vis-reasoning}
Wenbo Zheng, Lan Yan, Chao Gou, and Fei-Yue Wang. 2020.
\newblock Webly supervised knowledge embedding model for visual reasoning.
\newblock \emph{2020 IEEE/CVF Conference on Computer Vision and Pattern
  Recognition (CVPR)}, pages 12442--12451.

\end{thebibliography}
